\documentclass{article}

\usepackage{PRIMEarxiv}

\usepackage[utf8]{inputenc} 
\usepackage[T1]{fontenc}    
\usepackage{hyperref}       
\usepackage{url}            
\usepackage{booktabs}       
\usepackage{amsfonts}       
\usepackage{nicefrac}       
\usepackage{microtype}      
\usepackage{lipsum}
\usepackage{fancyhdr}       
\usepackage{graphicx}       
\graphicspath{{media/}}     

\usepackage{rotating}
\usepackage{ifpdf}
\usepackage[T1]{fontenc}
\usepackage{times}
\usepackage{sourcesanspro}
\usepackage{newtxmath}
\usepackage{textcomp}%
\usepackage{xcolor}%
\usepackage{hyperref}
\usepackage{appendix}
\usepackage{url}
\usepackage{xcolor}
\usepackage{tikz}

\usepackage[outdir=./]{epstopdf}

\usepackage{framed,multirow}

\usepackage{latexsym}

\usepackage{url}
\usepackage{xcolor}
\definecolor{newcolor}{rgb}{.8,.349,.1}

\usepackage{physics}  
\usepackage{empheq}
\usepackage{siunitx}
\usepackage{subcaption}

\usepackage{tikz}
\usepackage{xcolor}
\definecolor{blue}{RGB}{38,139,210}
\definecolor{cyan}{RGB}{42,161,152}
\definecolor{red}{RGB}{210, 90, 90}
\definecolor{base01}{RGB}{88,110,117}
\definecolor{base02}{RGB}{7,54,66}
\definecolor{base03}{RGB}{0,43,54}
\definecolor{base04}{RGB}{145,160,170}
\usetikzlibrary{calc,shapes,positioning}

\newcommand{\green}{G(\vb{x}, \vb{x}')}

\newcommand{\mrm}{\mathrm}
\newcommand{\U}{\vb{U}}

\newcommand{\x}{\vb{x}}
\newcommand{\vr}{\vb{r}}

\newcommand{\uv}{\vb{u}}

\DeclareMathSymbol{\upD}{\mathalpha}{operators}{1}
\newcommand{\veps}{\varepsilon}

\newcommand{\phiout}{\phi_\mrm{out}}
\newcommand{\Rin}{R_\mrm{in}}

\title{Using neural networks to solve the 2D Poisson equation for electric field computation in plasma fluid simulations}

\author{
  Lionel Cheng\\
  CERFACS \\
  42 Avenue Gaspard Coriolis\\
  31057 Toulouse, France\\
  \texttt{cheng@cerfacs.fr} \\
  \And
  Ekhi Ajuria Illarramendi\\
  ISAE-SUPAERO / CERFACS \\
  Université de Toulouse\\
  Toulouse, France\\
  \texttt{ajuria@cerfacs.fr} \\
  \And
  Guillaume Bogopolsky\\
  SAFRAN AIRCRAFT ENGINES / CERFACS \\
  Campus de l'Espace, 1 avenue Hubert Curien,\\
  27207 Vernon, France\\
  \texttt{bogopolsky@cerfacs.fr} \\
   \And
  Michaël Bauerheim\\
  ISAE-SUPAERO \\
  Université de Toulouse\\
  Toulouse, France\\
  \texttt{michael.bauerheim@isae-supaero.fr} \\
  \And
  Bénédicte Cuenot \\
  CERFACS \\
  42 Avenue Gaspard Coriolis\\
  31057 Toulouse, France\\
  \texttt{cuenot@cerfacs.fr} \\
}

\begin{document}
\maketitle

\begin{abstract}
The Poisson equation is critical to get a self-consistent solution in plasma fluid simulations used for Hall effect thrusters and streamer discharges, since the Poisson solution appears as a source term of the unsteady nonlinear flow equations. Two types of plasma fluid simulations are considered in this work: the canonical electron plasma oscillation and a double headed streamer discharge. As a first step, solving the 2D Poisson equation with zero Dirichlet boundary conditions using a deep neural network is investigated using multiple-scale architectures, defined in terms of number of branches, depth and receptive field \footnote{All the code in this work is available at \url{https://gitlab.com/cerfacs/plasmanet}}. One key objective is to better understand how neural networks learn the Poisson solutions and provide guidelines to achieve optimal network configurations, especially when coupled to the time-varying Euler equations with plasma source terms. Here, the Receptive Field is found critical to correctly capture large topological structures of the field. The investigation of multiple architectures, losses, and hyperparameters provides an optimal network to solve accurately the steady Poisson problem. The performance of the optimal neural network solver, called PlasmaNet, is then monitored on meshes with increasing number of nodes, and compared with classical parallel linear solvers. Next, PlasmaNet is coupled with an unsteady Euler plasma fluid equations solver in the context of the electron plasma oscillation test case. In this time-evolving problem, a physical loss is necessary to produce a stable simulation. PlasmaNet is finally tested on a more complex case of discharge propagation involving chemistry and advection. The guidelines established in previous sections are applied to build the CNN to solve the same Poisson equation in cylindrical coordinates with different boundary conditions. Results reveal good CNN predictions and pave the way to new computational strategies using modern GPU-based hardware to predict unsteady problems involving a Poisson equation, including configurations with coupled multiphysics interactions such as in plasma flows. 
\end{abstract}

\section{Introduction}
The Poisson equation is a well-known equation encountered in many fields of physics, from gravitation to incompressible flows, as well as in plasmas. In the context of plasma numerical simulations, the resolution of the Poisson equation for the electric field, which appears as source terms coupled to the nonlinear flow equations, is critical to properly describe the coupling with the charged particles' density fields. In this context, the Poisson equation relates the electromagnetic potential $\phi$ to the charge distribution $\rho_q$:

\begin{equation}
    \laplacian \phi = -\frac{\rho_q}{\varepsilon_0}
    \label{eq:poisson_eq}
\end{equation}
\noindent where $\veps_0$ is the magnetic permeability in vacuum.

The potential gives access to the electric field as $\vb{E} = - \grad \phi$. Plasma fluids are made of transport charged species $i$ of density $n_i$, velocity $\vb{v}_i$ and specific energy $E_i$ governed by the Euler transport equations in which the electric field appears as a source term. The classical way to solve the Poisson equation is to discretize the Laplace operator on a mesh so that Eq.~\eqref{eq:poisson_eq} reduces to a linear system $A \phi = R$, where $A$ is the Laplace operator matrix and $R$ is the discretized version of the charge density. Direct or iterative methods can be used to solve such linear systems \cite{quarteroni}. The computational cost of solving Eq.~\eqref{eq:poisson_eq} increases with the mesh size. Following the idea of FluidNet \cite{tompson2017accelerating, ajuria2020towards}, this work introduces data-driven methods to accelerate the resolution of the Poisson equation, and investigate their behavior when coupled to unsteady Euler equations.

The resolution of Partial Differential Equations (PDE) using Machine Learning techniques was first developed in the mid-1990s, with the introduction of MultiLayer Perceptrons (MLP)~\cite{rosenblatt1958perceptron} to solve a 2D Poisson equation with Dirichlet boundary conditions~\cite{dissanayake1994neural}. Although showing promising results, these first attempts were quickly limited by the available computational resources.  With today's computational power as well as the recent development of user-friendly Machine Learning frameworks, the interest in the resolution of PDEs with Machine Learning has significantly grown. One of the main milestones in the field was the introduction of \textit{Physics informed neural networks} (PINN)~\cite{raissi2017physics} which incorporates physical knowledge into the neural networks. For example, training the network to minimize the residual of physical equations imposes a physical constraint on the network. Such networks employ automatic differentiation~\cite{baydin2018automatic} resulting in mesh-free methods. Yet, PINNs have to be trained specifically for each boundary or initial condition, which limits their practical use, in particular in Computational Fluid Dynamics (CFD).

The capability of neural networks to approximate the solution of PDEs has thus led to the development of various surrogate models for fluid mechanics simulation. For example, Convolutional Neural Networks (CNN) trained with physical loss functions have been used to substitute incompressible fluid solvers~\cite{wandel2020unsupervised,tompson2017accelerating, ajuria2020towards}), obtaining stable simulations with considerable speed-up. An \textit{a posteriori} physical correction of the network predictions have also been reported successful in time-evolving problems \cite{alguacil2021predicting}, avoiding error accumulation in time. For a complete review of the use of Machine Learning in fluid dynamics, the reader is referred to the review of Brunton et al.~\cite{brunton2020machine}.

Instead of completely substituting a CFD solver, this work focuses on one particular step of the resolution process: the Poisson equation. Initial attempts using MLPs~\cite{yang2016data} were quickly followed by the introduction of CNNs~\cite{ozbay2021poisson, shan2020study}, which were better suited to map the input and output in 2 or 3 dimensions. Nevertheless, they still treated the network as a separate instance to the fluid solver, as they were trained \textit{outside the box} and were not coupled to a CFD solver to validate the methodology on steady or unsteady simulations.

Recent works have used CNNs~\cite{tompson2017accelerating} to solve the Poisson equation, coupled to the fluid solver, embedding the concept of a simulation into the neural network. Additionally, the hybridization of such embedded networks with standard Poisson solvers provided fast and robust CFD solvers, especially on configurations with new physics (for example a network trained with constant density flow, but tested on variable density simulations \cite{ajuria2020towards}). Similarly, differentiable fluid solvers~\cite{um2020solver} have also recently been introduced, which enabled to encode flow dynamics during the training process.

These works are extended here by training a deep CNN to solve the Poisson equation in the context of plasma flows, where it is used to obtain the electric field from a charge distribution. A first objective is to better understand how to design the neural architecture and its associated hyperparameters to achieve stable and accurate plasma flow simulations. A second objective is therefore to couple the data-driven Poisson solver to a multiphysics Euler plasma unsteady solver, and evaluate the resulting accuracy and performance (accuracy referring to the precision of the network prediction and performance to the network inference time). In such time-evolving multiphysics problems, it is found critical (1) to train the network using a physical-based loss function, and (2) to design optimal network architectures, for which the receptive field is found essential. In section 2, a first analytical test problem is presented. The methodology based on CNN is described in section 3 with different loss functions and architectures. In section 4, the datasets used for training and validation are presented. The accuracy and performance of the neural network solver is assessed in section 5. In section 6 the coupling of the neural network solver with an Euler plasma flow solver is shown in a canonical test case. Finally, in section 7, a more complex test case of plasma discharge propagation in a cylindrical geometry is tackled using the experience gained from the previous study in cartesian geometry.

\section{Problem configuration and solution}

The objective is to test the network-based Poisson solver on an academic plasma-fluid problem where analytical solutions exist. To do so, a simple unsteady problem on a rectangular domain is investigated, with boundary conditions and geometry allowing analytical resolution. Note however that the present methodology, as FluidNet~\cite{tompson2017accelerating} for incompressible flows, is not restricted to rectangular domains and simple geometries.

Plasmas are composed of charged species that can be modeled in a fluid formulation \cite{bittencourt}. In its simplest form, each charged species $i$ has its own set of Euler equations with electromagnetic source terms (reduced only to the electric field here) and no collision source terms (supposed to be negligible here by assuming a low-temperature as done in \cite[Chap. 11.1]{bittencourt}):
\begin{align}
\pdv{\U_i}{t} + \div \vb{F}_i &= \vb{S}_i \qq{for all species}i \\
\nabla\cdot\mathbf{E} &= \frac{\rho_q}{\varepsilon_0} \label{eq:maxwell_gauss}
\end{align}
where
\begin{align}
\U_i = \begin{bmatrix}
\rho_i \\ \rho_i \uv_i \\ \rho_i E_i
\end{bmatrix}
\quad
\vb{F}_i = \begin{bmatrix}
\rho_i \uv_i \\ \rho_i \uv_i \otimes \uv_i + p_i \vb{I} \\ (\rho_i E_i + p_i)\uv_i
\end{bmatrix}
\qq{and}
\vb{S}_i = \begin{bmatrix}
    0 \\ q_i \, n_i \, \vb{E} \\ q_i \, n_i \, \vb{E} \cdot \uv_i
\end{bmatrix}
\label{eq:plasma_transport}
\end{align}
where $\rho_i$ is the mass density, $n_i$ the number density, $\uv_i$ the speed, $E_i$ the total energy per mass unit, $p_i$ the pressure of species $i$. Finally $\rho_q = \sum q_i \, n_i$ is the charge density and $\vb{E}$ the electric field. From now on, $R = \rho_q / \veps_0$ is set to alleviate notations. It will be abusively referred to as the charge density althrough strictly speaking the charge density is $\rho_q$.

In this set of equations, the charged species interact with one another to yield an electric field through the Maxwell-Gauss equation \eqref{eq:maxwell_gauss}. Without magnetic field, the electric field can be written as $\vb{E} = - \nabla \phi$ and so Eq.~\eqref{eq:maxwell_gauss} becomes a Poisson equation \eqref{eq:poisson_eq}. When discretizing Eq.~\eqref{eq:poisson_eq}, the resulting linear system requires iterative methods such as Jacobi or Conjugate gradient \cite{quarteroni} to be solved. In most plasma-fluid simulations of real applications, this resolution represents up to 80\% of the total CPU cost and limits the computation capability.

Eq.~\eqref{eq:poisson_eq} must be completed with proper boundary conditions, taken here as zero-Dirichlet conditions, (\textit{i.e.}, imposing $\phi = 0$ on all the boundaries). The equation is computed on a square domain with uniform spacing $\Delta$ in the $x$ and $y$ directions. Thus the problem can be recast as

\begin{empheq}[left=\empheqlbrace]{align}
  \nabla^2 \phi &= - R \qq{in} \mathring{\Omega} \label{eq:poisson_interior} \\
    \phi &= 0 \qq{on} \partial \Omega \label{eq:dirichlet_bc}
\end{empheq}

\noindent where $\mathring{\Omega}$ refers to the internal computational domain and $\partial \Omega$ to its boundary. Applying Dirichlet boundary conditions on a domain of size $(L_x, L_y)$ allows the derivation of analytical solutions, written in terms of spatial Fourier series (\ref{appendix:analytical_solution}) as

\begin{align}
    \phi(x, y) &=  \sum_{n = 1}^{+\infty}\sum_{m = 1}^{+\infty} \phi_{nm} \sin\qty(\frac{n\pi x}{L_x})\sin\qty(\frac{m\pi y}{L_y}) \label{eq:analytical_solution} \\
    \phi_{nm} &= \frac{R_{nm}}{\qty[\qty(\frac{n\pi}{L_x})^2 + \qty(\frac{m\pi}{L_y})^2]} \label{eq:pot_coefficients} \\
    R_{nm} &=  \frac{4}{L_x L_y} \int_{x', y'} \sin\qty(\frac{n\pi x'}{L_x})\sin\qty(\frac{m\pi y'}{L_y})R(x', y') \dd{x'} \dd{y'}  \label{eq:R_coefficients}
\end{align}

\noindent where the profile of charge density $R(x', y')$ may be of any kind.

\begin{figure}[htbp]
    \centering
    \includegraphics[width = \textwidth]{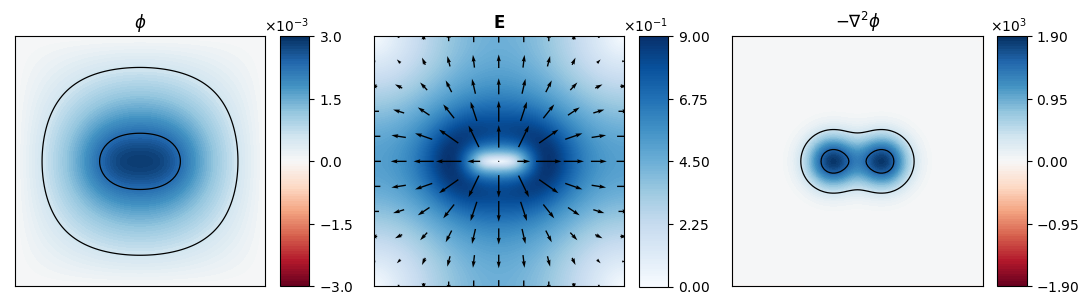}
    \caption{Potential [V] and electric field [V.m$^{-1}$] (norm in color and direction with arrows) associated to a two-Gaussians charge density [V.m$^{-2}$] in a 1 cm$^2$ domain.}
    \label{fig:problem_example}
\end{figure}

To compute the analytical solution, spatial Fourier modes $R_{nm}$ of the charge density are first computed Eq.~\eqref{eq:R_coefficients}. Then the potential Fourier coefficients $\phi_{nm}$ are derived from $R_{nm}$ with Eq.~\eqref{eq:pot_coefficients}. Thus the resulting potential has a diffused shape compared to the charge density. An example is given in Fig.~\ref{fig:problem_example}, showing the potential ($\phi$, left) and the electric field ($\vb{E}$, middle), obtained with a charge distribution ($R = - \nabla^2 \phi$, right) consisting of two Gaussian functions. The two Gaussian peaks, clearly visible on the charge density field, are totally diffused and merged in the resulting potential, highlighting the low-pass filter behavior of the inverse Laplacian operator. Note that this solution is of little practical use: for $N$ (resp. $M$) modes along the $x$ (resp. $y$), $N\times M$ integral evaluations Eq.~\eqref{eq:analytical_solution} are necessary to compute the potential. For a low-frequency charge density profile such as the two gaussians, $N = M = 10$ modes in each direction are required to get below 1\% of error on the 1-norm of the electric field. For high-frequency profiles such as the one depicted in Fig.~\ref{fig:random_8_dataset} $N = M = 50$ modes in each direction is necessary so that 2500 domain integrals need to be computed. Therefore this solution can help us understand the structure of the solution and can serve as a reference solution for comparison with neural network predictions, but it is too computationally expensive to be used in a simulation.

\section{Neural networks and methods}
Deep neural networks are composed of multiple tunable neurons that can learn complex functions. To do so, a non-convex optimization procedure is performed to update the neuron weights by minimizing a cost function (the loss function). This loss function is crucial, and several metrics are presented and tested in this work. Neural networks are denoted by $f$ such that

\begin{equation}
    \phi_\text{out} = f(R_\text{in})
\end{equation}

\noindent The target solution (potential obtained with a linear solver) is on the other hand denoted by $\phi_\text{target}$.

\subsection{Loss functions}

Two kinds of loss functions are used: one for the interior points corresponding to $\mathring{\Omega}$ and the other one for the boundary points $\partial \Omega$, as introduced in Eqs.~\eqref{eq:poisson_interior}-\eqref{eq:dirichlet_bc}.

In deep neural networks on supervised tasks, pixel-to-pixel distances are often used as loss functions for training and are defined as \texttt{InsideLoss} Eq.~\eqref{eq:insideloss} and \texttt{DirichletLoss} Eq.~\eqref{eq:dirichletloss}. For the \texttt{DirichletLoss}, the target value of the boundary points is known from the problem definition (0 in the chosen configuration here). However for the \texttt{InsideLoss} a pre-computed target dataset to which the network prediction is compared is needed. An alternative is the \texttt{LaplacianLoss} which uses the residual of Eq.~\eqref{eq:poisson_interior} and therefore avoids to solve the Poisson equation with linear solvers. Two combinations of losses will be thus tested: \texttt{DirichletLoss} - \texttt{InsideLoss}~and~\texttt{DirichletLoss} - \texttt{LaplacianLoss}.

\begin{itemize}
    \item \texttt{DirichletLoss}:
    \begin{equation}
        \label{eq:dirichletloss}
        \mathcal{L}_D(\vb{\phi}_\text{out}) = \frac{1}{b_s (2n_x + 2n_y - 4)} \sum_{b, j, i} (\phi_\text{out}^{b, j, i})^2
    \end{equation}
    \item \texttt{InsideLoss}:
    \begin{equation}
        \label{eq:insideloss}
        \mathcal{L}_I(\vb{\phi}_\text{out}, \vb{\phi}_\text{target}) = \frac{1}{b_s (n_x - 1) (n_y - 1)} \sum_{b, j, i} \qty(\phi_\text{out}^{b, j, i} - \phi_\text{target}^{b, j, i})^2
    \end{equation}
    \item \texttt{LaplacianLoss}:
    \begin{equation}
        \label{eq:laplloss}
        \mathcal{L}_L(\vb{\phi}_\text{out}) = \frac{L_x^2 L_y^2}{b_s (n_x - 1) (n_y - 1)} \sum_{b, j, i} \qty(\nabla^2\phi_\text{out}^{b, j, i} + R_\text{in}^{b, j, i})^2
    \end{equation}
\end{itemize}

\noindent where $b_s$, $n_x$, $n_y$ refer to the batch size (number of training examples used in one iteration of learning), number of nodes in $x$ and $y$ directions, $b, j, i$ refer to the indices of the batch size, the $y$ and $x$ directions respectively.

\subsection{Network architectures}
The analytical derivation Eq.~\eqref{eq:analytical_solution} highlighted the low-pass behavior of the inverse Laplace operator. Spatial scales are therefore crucial when solving a Poisson equation, which therefore drives the choice of the neural network: the Multi-Scale (denoted MSNet) architecture \cite{multi_scale_paper} and UNet architecture \cite{ronneberger2015u} embed the notion of spatial scales, with dedicated treatment of the various scales contained in the input.

\begin{figure}[htbp]
    \centering
    \includegraphics[width = 0.8\textwidth]{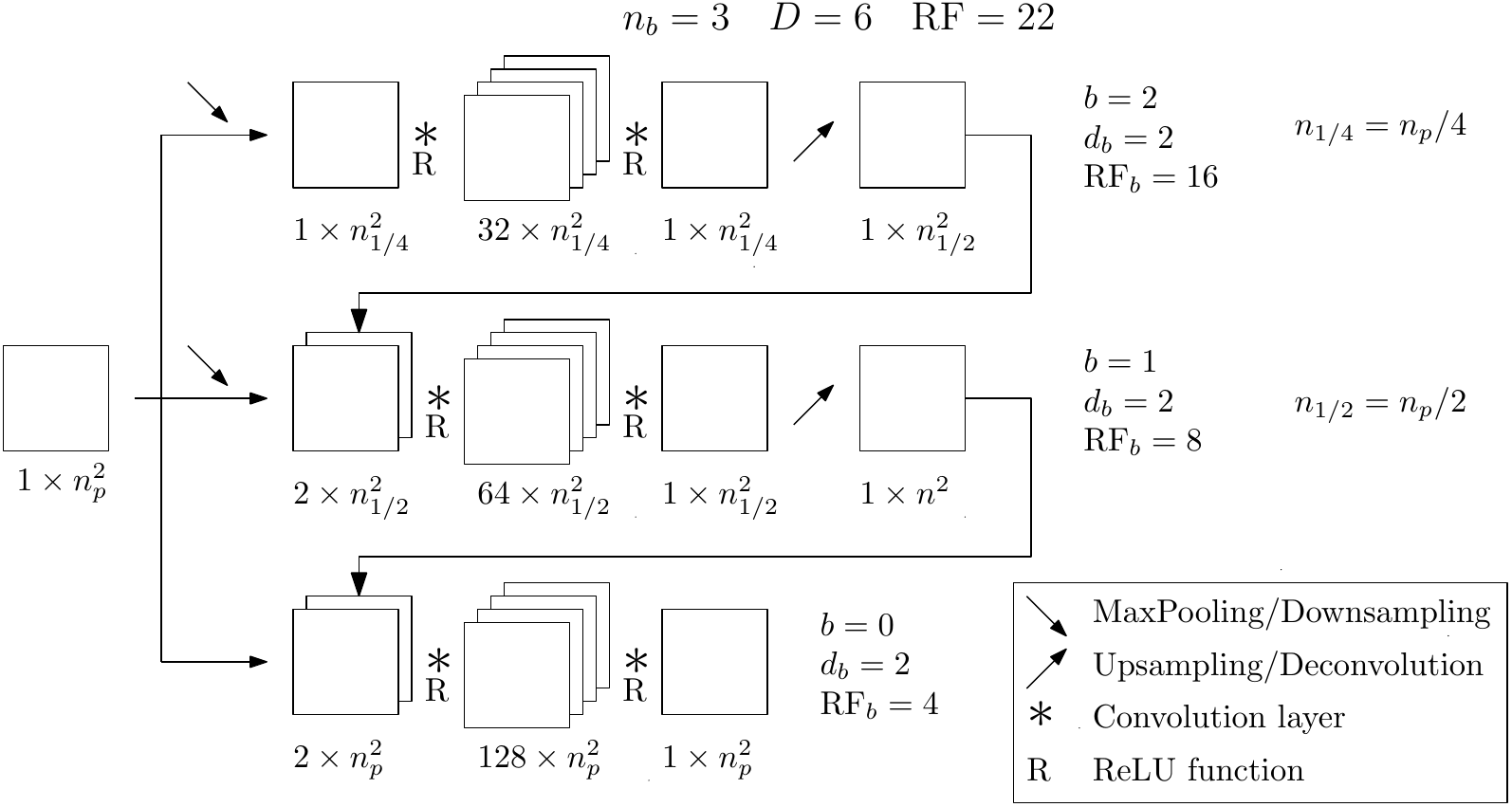}
    \caption{Sketch of MSNet with $n_s$ = 3.}
    \label{fig:msnet_sketch}
\end{figure}

The MSNet has been designed for video predictions \cite{multi_scale_paper} for which classical deep networks failed to capture accurately the largest scales of the inputs. This architecture has already been used for different flow applications, such as the super-resolution of turbulent flows~\cite{fukami2019super} and the propagation of acoustic waves~\cite{alguacil2020predicting}. Generally speaking this network consists of $n_s$ scale channels (a sketch of a $n_s=3$ network is presented in Fig.~\ref{fig:msnet_sketch}). In each channel $i$, the initial images are downsampled from $n_p$ pixels to $\left \lfloor{n_p / 2^i}\right \rfloor$ pixels per direction. From now on the image resolution is either defined in terms of spatial spacing $\Delta$ or number of pixels ($n_p$). A series of convolutional layers is applied and the output of channel $i$ goes to channel $i - 1$  except for the $i = 0$ scale channel.

\begin{figure}[htbp]
  \centering
  \includegraphics[width = 0.8\textwidth]{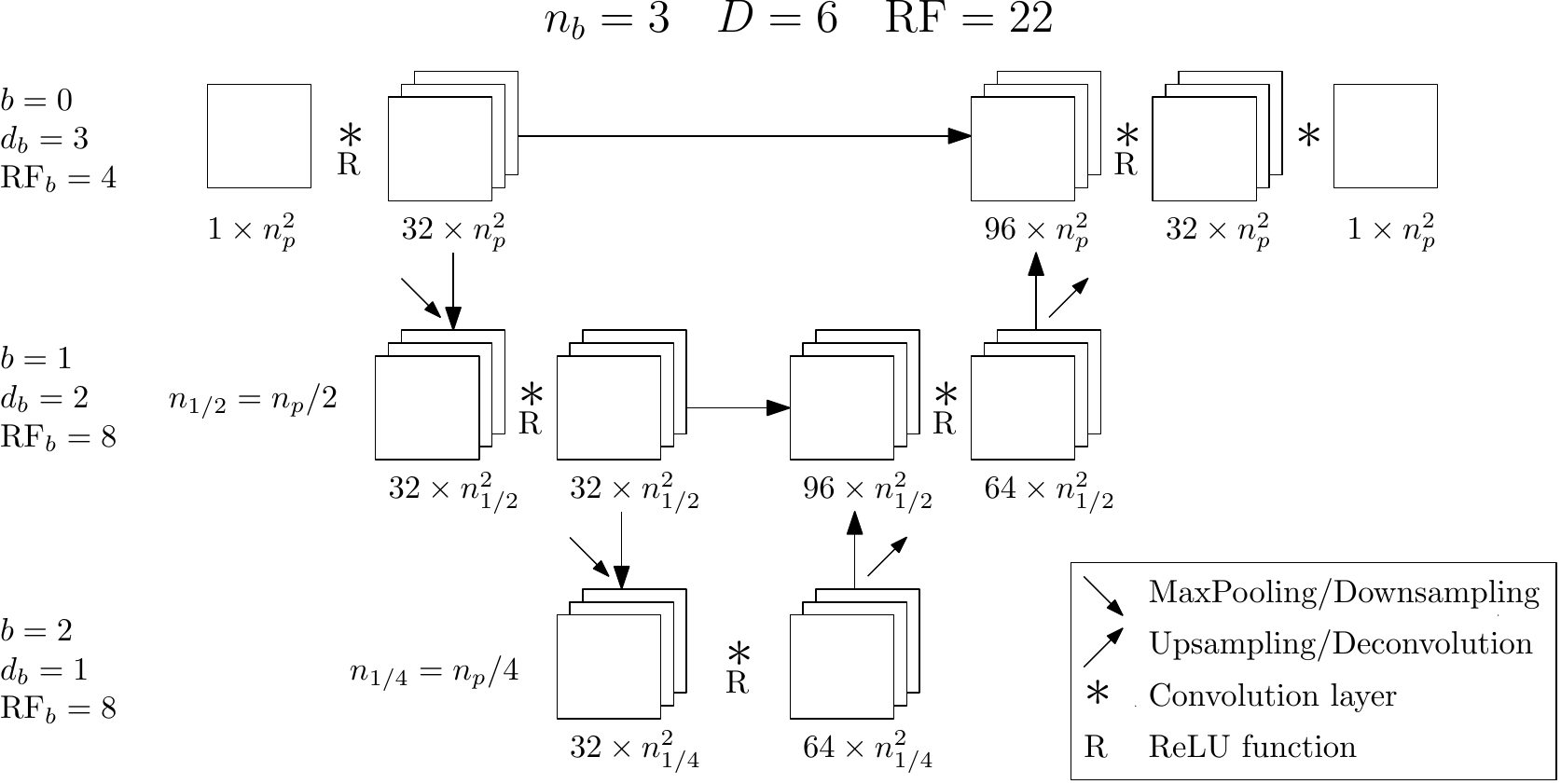}
  \caption{Sketch of UNet with $n_s = 3$.}
  \label{fig:unet_sketch}
\end{figure}

The UNet was first introduced for biomedical segmentation \cite{ronneberger2015u}, and it has been widely used by the ML-CFD community~\cite{lapeyre2019training, thuerey2020deep}. A sketch is presented in Fig.~\ref{fig:unet_sketch}. A series of encoding layers are applied, each time decreasing the number of pixels by a power of 2 in each direction as in the MSNet. What differentiates the UNet from a simple encoder-decoder network is the skip connection which links every enconding layer to its decoding counterpart.

In both MSNet and UNet, the local branch $b$ is defined as the power of 2 by which the initial resolution is divided by, in a specific scale of the network. The number of branches $n_b$ of the network is the number of scales involved ($n_b = n_s$).

The local depth $d_b$ is defined as the number of successive convolutional layers in branch $b$. The global depth $D$ of the network is then the sum of the local depths, assuming that $D$ corresponds to the longest path from input to output. The higher the depth of the network, the more likely vanishing or exploding gradients can appear.

\begin{equation}
D = \sum_{b=0}^{n_b - 1} d_b
\end{equation}

\subsubsection{Receptive field}

\begin{figure}[htbp]
    \centering
    \begin{subfigure}[b]{0.45\textwidth}
        \begin{tikzpicture}[scale=.5,every node/.style={minimum size=1cm},on grid]
    \begin{scope}[xshift=0, yshift=0, every node/.append style={yslant=0,xslant=0},
                    yslant=0.5,xslant=-0.9]
        \draw[fill=base01] (2, 2) rectangle
        (2.5, 2.5);
        \draw[fill=base01, opacity=0.4] (0, 0) rectangle
                                (5, 5);

        \draw[step=5mm, base03, thin] (0, 0) grid
                                        (5, 5);
        \draw[base03, thick] (0, 0) rectangle (5, 5);
        
        \coordinate (BL0) at (2, 2);
        \coordinate (BR0) at (2.5, 2);
        \coordinate (TL0) at (2, 2.5);
        \coordinate (TR0) at (2.5, 2.5);
        \node at (6, -1) {Layer 0};
    \end{scope}

    \begin{scope}[xshift=-5, yshift=110,
                    every node/.append style={yslant=0,xslant=0},
                    yslant=0.5,xslant=-0.9]
        \node at (6, -1) {Layer 1};

        \coordinate (BL1) at (1.5, 1.5);
        \coordinate (BR1) at (3, 1.5);
        \coordinate (TL1) at (1.5, 3);
        \coordinate (TR1) at (3, 3);

        \draw (BL0) -- (BL1) (BR0) -- (BR1)
              (TL0) -- (TL1)  (TR0) -- (TR1);
        
        \draw[fill=red] (1.5, 1.5) rectangle (3, 3);
        \draw[fill=red, opacity=0.4] (0, 0) rectangle (5, 5);
        \draw[step=5mm, base03, thin] (0,0) grid (5, 5);
        \draw[base03, thick] (0, 0) rectangle (5, 5);
    \end{scope}

    \begin{scope}[xshift=-5, yshift=220,
        every node/.append style={yslant=0,xslant=0},
        yslant=0.5,xslant=-0.9]
        \node at (6, -1) {Layer 2};

        \coordinate (BL2) at (1, 1);
        \coordinate (BR2) at (3.5, 1);
        \coordinate (TL2) at (1, 3.5);
        \coordinate (TR2) at (3.5, 3.5);

        \draw (BL1) -- (BL2) (BR1) -- (BR2)
        (TL1) -- (TL2)  (TR1) -- (TR2);

        \draw[fill=blue] (1, 1) rectangle (3.5, 3.5);
        \draw[fill=blue, opacity=0.4] (0, 0) rectangle (5, 5);
        \draw[step=5mm, base03, thin] (0,0) grid (5, 5);
        \draw[base03, thick] (0, 0) rectangle (5, 5);
    \end{scope}
\end{tikzpicture}
        \caption{$b = 0$ \quad $d_b = 2$ \quad $k_s = 3$}
        \label{fig:rf_h0}
    \end{subfigure}
    \begin{subfigure}[b]{0.45\textwidth}
        \begin{tikzpicture}[scale=.5,every node/.style={minimum size=1cm},on grid]
    \begin{scope}[xshift=0, yshift=0, every node/.append style={yslant=0,xslant=0},
                    yslant=0.5,xslant=-0.9]
        \draw[fill=base01] (2, 2) rectangle
        (3, 3);
        \draw[fill=base01, opacity=0.4] (0, 0) rectangle
                                (5, 5);

        \draw[step=5mm, base04, thin] (0, 0) grid
                                        (5, 5);
        \draw[step=1cm, base03, thick] (0,0) grid (5, 5);
        
        \coordinate (BL0) at (2, 2);
        \coordinate (BR0) at (3, 2);
        \coordinate (TL0) at (2, 3);
        \coordinate (TR0) at (3, 3);
        \node at (6, -1) {Layer 0};
    \end{scope}

    \begin{scope}[xshift=-5, yshift=110,
                    every node/.append style={yslant=0,xslant=0},
                    yslant=0.5,xslant=-0.9]
        \node at (6, -1) {Layer 1};

        \coordinate (BL1) at (1, 1);
        \coordinate (BR1) at (4, 1);
        \coordinate (TL1) at (1, 4);
        \coordinate (TR1) at (4, 4);

        \draw (BL0) -- (BL1) (BR0) -- (BR1)
              (TL0) -- (TL1)  (TR0) -- (TR1);
        
        \draw[fill=red] (1, 1) rectangle (4, 4);
        \draw[fill=red, opacity=0.4] (0, 0) rectangle (5, 5);
        \draw[step=5mm, base04, thin] (0,0) grid (5, 5);
        \draw[step=1cm, base03, thick] (0,0) grid (5, 5);
        \draw[base03, thick] (0, 0) rectangle (5, 5);
    \end{scope}

    \begin{scope}[xshift=-5, yshift=220,
        every node/.append style={yslant=0,xslant=0},
        yslant=0.5,xslant=-0.9]
        \node at (6, -1) {Layer 2};

        \coordinate (BL2) at (0, 0);
        \coordinate (BR2) at (5, 0);
        \coordinate (TL2) at (0, 5);
        \coordinate (TR2) at (5, 5);

        \draw (BL1) -- (BL2) (BR1) -- (BR2)
        (TL1) -- (TL2)  (TR1) -- (TR2);

        \draw[fill=blue] (0,0) rectangle (5, 5);
        \draw[step=5mm, base04, thin] (0,0) grid (5, 5);
        \draw[step=1cm, base03, thick] (0,0) grid (5, 5);
        \draw[base03, thick] (0, 0) rectangle (5, 5);
    \end{scope}
\end{tikzpicture}
        \caption{$b = 1$ \quad $d_b = 2$ \quad $k_s = 3$}
        \label{fig:rf_h1}
    \end{subfigure}
    \caption{Domain of influence of the center point across two convolutional layers with a kernel size of 3.}
    \label{fig:receptive_field_sketch}
\end{figure}
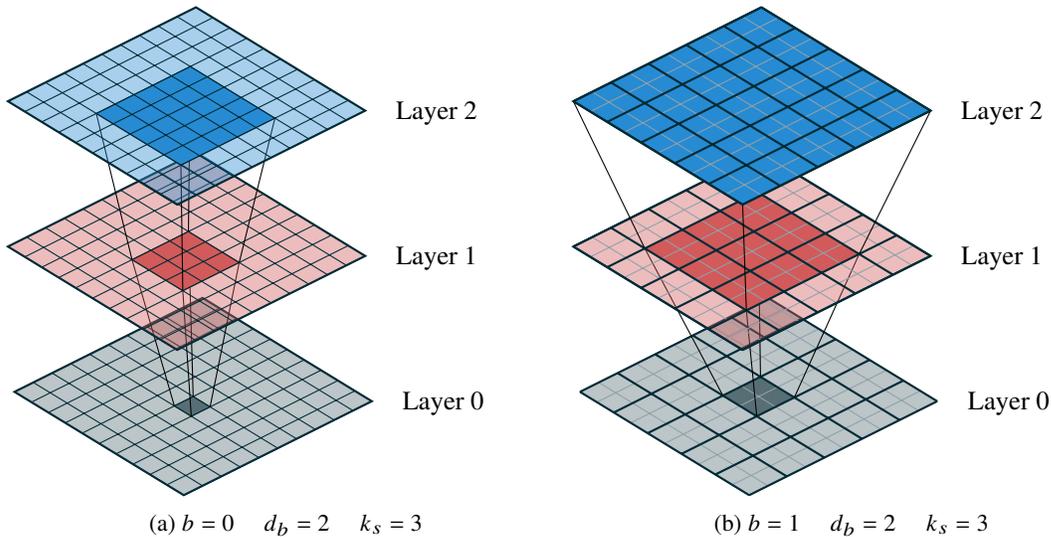

In a convolutional neural network (CNN), from one layer to the next, information propagates according to the kernel size $k_s$ of the convolution layer. Typically, one layer with $k_s = 3$ gathers information from a neighboring of $3 \times 3$ pixels only, and therefore cannot capture large structures. However, when several layers of convolution are applied, points that are far away from each other in the input image can interact with each other, which allows larger structures to be detected. As an illustration, the domain of influence of the center point of an image with two convolutional layers at different branches $b$ is depicted in Fig.~\ref{fig:receptive_field_sketch}. To help capturing the largest scales, MSNet and UNet employ downscaled branches: by downscaling the input image by a factor 2, the second branch in Fig.~\ref{fig:rf_h1} tackles flow structures twice as big as the first branch in Fig.~\ref{fig:rf_h0}.

To quantify the information propagation the global receptive field $\mrm{RF}$ is defined as the size of the domain of influence of the input center point in number of points in the original scale $n_p$. The receptive field can be splitted into local receptive fields per branch $\mrm{RF}_b$ so that

\begin{equation}
\label{eq:rf_def}
\mrm{RF} = \sum_{b=0}^{n_b - 1} \mrm{RF}_b
\end{equation}
with
\begin{equation}
\mrm{RF}_b =
\begin{cases}
    1 + d_b (k_s - 1) 2^b \quad \text{if} \ b = 0\\
    d_b (k_s - 1) 2^b \qq{otherwise}
\end{cases}
\end{equation}

\noindent where the branch $b =0$ needs to take into account the original input pixel (+1) and $k_s$ is the kernel size of all the convolutional layers of the network (supposed to be equal). In Fig.~\ref{fig:rf_h0}, $b=0, k_s = 3$ so that every layer extends the receptive field by $(3 - 1)2^0 = 2$ and the branch receptive field is $\mrm{RF}_0 = 1 + 4 = 5$. In Fig.~\ref{fig:rf_h1}, $b=1, k_s=3$, the initial $\mrm{RF}$ from previous branches is equal to 2 ($\mrm{RF}_0 = 1$) and the grid that is actually processed by the convolutional kernel is 2 times coarser (black thick line) compared to the orignial grid (thin gray line). Hence every layer extends the receptive field by $(3 - 1)2^1 = 4$ which is 2 times more than the previous branch yielding a receptive field of $\mrm{RF}_1 = 8$ for that branch and the total receptive field in the end is $\mrm{RF} = \mrm{RF}_0 + \mrm{RF}_1 = 10$ which corresponds to the size of layer 2.

In \cite{chua1993cnn}, a \textit{theoretical receptive field} is defined as the size of the input domain of influence on the output center point. Tests carried on the studied UNets and MSNets show that this definition matches Eq.~\eqref{eq:rf_def} so that both formulations are equivalent. However, the importance of the domain influence is not uniform, as points closer to the studied pixel will have more paths to influence the output, resulting in a \textit{gaussian-like} distribution~\cite{luo2016understanding}. The distribution helps to understand the behavior of the CNN, as it will especially focus on a smaller centered region, known as \textit{effective receptive field}, while still being influenced by information located on the boundaries of the~\textit{receptive field}.

One convolutional layer in branch $b$ contributes two times more to the receptive field than a convolutional layer in branch $b-1$. The local and global properties defined in this section are reported for the simple MSNet3 and UNet3 in Fig.~\ref{fig:unet_sketch}.

The receptive field analysis is particularly of interest in view of the elliptic nature of the Poisson equation, which does not load to characteristic lines and propagates information only spatially, every point of the domain influencing the whole domain. This suggests that the neural network should see the whole input image to predict correctly the solution.

\subsection{Normalization}

A neural network learns well from inputs and outputs that span approximately the same range of values~\cite{bishop1995neural}. For example in image prediction, the network needs to output a field of values in the interval [0 ,1] from values that are also in [0, 1]~\cite{chollet2017deep}. In the present case, the potential maximum value is not known \textit{a priori}, so that only the density charge is rescaled as

\begin{align}
    \phiout = f(\tilde{\Rin}) \qq{where} \tilde{\Rin} = \Rin \times \qty|\frac{\phiout}{\Rin}|_\text{max}
\end{align}

\noindent which ensures maximum values of the input and output of the same order.

A reasonable value for the ratio of the potential over the charge density needs to be found. From the solution of the potential in terms of Fourier series, the following normalization is chosen:

\begin{equation}
    \qty|\frac{\phiout}{\Rin}|_\text{max} = \frac{\alpha}{\qty(\frac{\pi^2}{4})^2\qty(\frac{1}{L_x^2} + \frac{1}{L_y^2})}
\end{equation}

\noindent where $\alpha = 0.1$ is used. More details on the derivation of this relation can be found in \ref{appendix:normalization}. This value is actually correlated with the FWHM (Full Width Half Maximum) of the charge density $R$. The higher the value of the FWHM, the higher the maximum value of the potential $\phi$.

Besides bringing the values of the input and output of the networks to the same orders of magnitude, this normalization also brings similarity in domain-length: the normalized solution in a square box of length $L_x$ and resolution $\upD$ is similar to the one in a square box of length $\alpha L_x$ and resolution $\alpha \upD$.

\subsection{Resolution scaling}

\label{subsec:res_scaling}

After the training of the network in a square box of length $L_x$ and resolution $\upD_1$ the problem of the applicability of this network to a square box of length $L_x$ and resolution $\upD_2$ is discussed in this section.

The Laplacian operator with a resolution $\upD$ can be nondimensionalized:

\begin{equation}
    \nabla^2_{\upD} = \pdv[2]{x} + \pdv[2]{y} = \frac{1}{\upD^2} \qty(\pdv[2]{\bar{x}} + \pdv[2]{\bar{y}})
\end{equation}

\noindent where the overbar indicates nondimensionalized physical values. The nondimensionalized operator should be the conserved quantity between two resolutions. Denoting by $\upD_\mrm{sim}$ the resolution of the simulation and by $\upD_\mrm{NN}$ the resolution at which the neural network was trained, the following relationship holds:

\begin{equation}
    \nabla^2_{\upD_\mrm{sim}} = \frac{\upD_\mrm{NN}^2}{\upD_\mrm{sim}^2} \nabla^2_{\upD_\mrm{NN}}.
\end{equation}
What the network is emulating is in fact the inverse of the Laplacian, hence:

\begin{equation}
    (\nabla^2_{\upD_\mrm{sim}})^{-1} = \frac{\upD_\mrm{sim}^2}{\upD_\mrm{NN}^2} (\nabla^2_{\upD_\mrm{NN}})^{-1}
    \label{eq:resolution_invariance}
\end{equation}

\noindent so that the initial neural network guess needs to be multiplied by a ratio of resolutions to be applied to other resolutions.

\section{Datasets}

In the present work, two types of datasets are proposed, where the spatial scales can be controlled. The objective is to better understand how the neural network can learn, and then predict accurately the various scales of the solution in the context of the Poisson equation. To do so, two types of datasets will be generated: (i) a random dataset, and (ii) a random Fourier dataset.

\subsection{Random dataset}

\begin{figure}[htbp]
    \centering
    \includegraphics[width=0.55\textwidth]{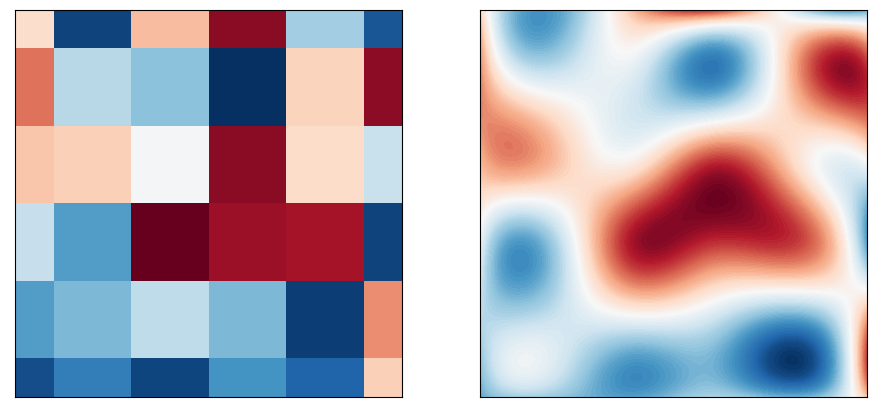}
    \caption{Random values taken in a $6 \times 6$ coarse grid (left) and interpolated in a $101 \times 101$ fine grid (right) where $c = 16$.}
    \label{fig:random_generation}
\end{figure}

First proposed by Ozbay \cite{ozbay2021poisson}, a random distribution of values in the range $[-1 ,1]$ is generated in a coarse grid of size $n_\mrm{coarse} = \lfloor n_p / c \rfloor$, $n_p$ being the number of points in each direction and $c$ a chosen filter size. Then bicubic interpolation generates a random field with controlled structure size on the target grid. The minimum structure is of size $c$ pixels. This procedure is illustrated for $c=16$ in Fig.~\ref{fig:random_generation}. From now on, \texttt{random\_c} will be used to denote $c$-random datasets.

\begin{figure}[htbp]
    \centering
    \includegraphics[width=\textwidth]{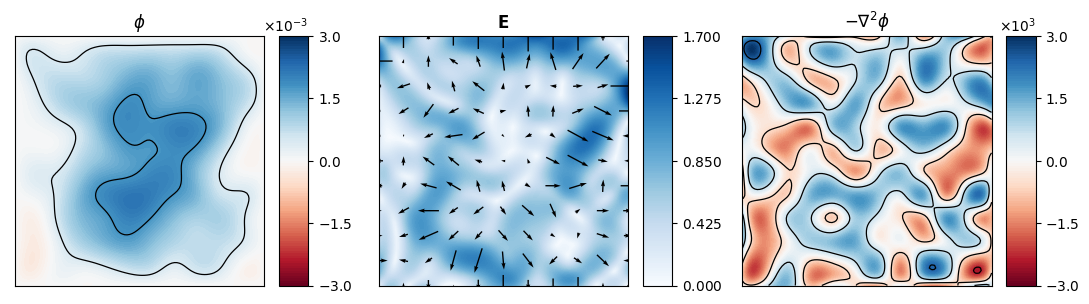}
    \caption{Example of \texttt{random\_8} source term input in a 1 cm$^2$ domain.}
    \label{fig:random_8_dataset}
\end{figure}

An example of $(\phi, \vb{E}, R = - \nabla^2 \phi)$ is shown in Fig.~\ref{fig:random_8_dataset} for a random dataset filetered with $c=8$.

\subsection{Random Fourier dataset}

Also proposed by Ozbay \cite{ozbay2021poisson}, a random Fourier dataset is obtained by setting randomly the coefficients $R_{nm}$ of Eq.~\eqref{eq:R_coefficients}, and introducing maximum frequencies ($N$ and $M$) in the sums:

\begin{equation}
    A(x, y) = \sum_{n = 1}^{N} \sum_{n = 1}^{M} A_{nm} \sin(\frac{n\pi x}{L_x}) \sin(\frac{m\pi y}{L_y}) \qq{for} A \in \{\phi, R\}
    \label{eq:natural_modes}
\end{equation}

\noindent where $\phi_{nm}$ is deduced from $R_{nm}$ thanks to Eq.~\eqref{eq:pot_coefficients}.

The value of $R_{nm}$ are taken randomly following a power $p$ decreasing law to mimic the high-frequency damping of physical solutions:

\begin{equation}
    R_{nm}(p) \sim \frac{1}{n^p + m^p} \frac{en_0}{\varepsilon_0} \, \mathcal{U}(-1, 1).
\end{equation}

\noindent where $\mathcal{U}(-1, 1)$ corresponds to a uniform distribution over the range $[-1, 1]$.

\begin{figure}[htbp]
    \centering
    \includegraphics[width=\textwidth]{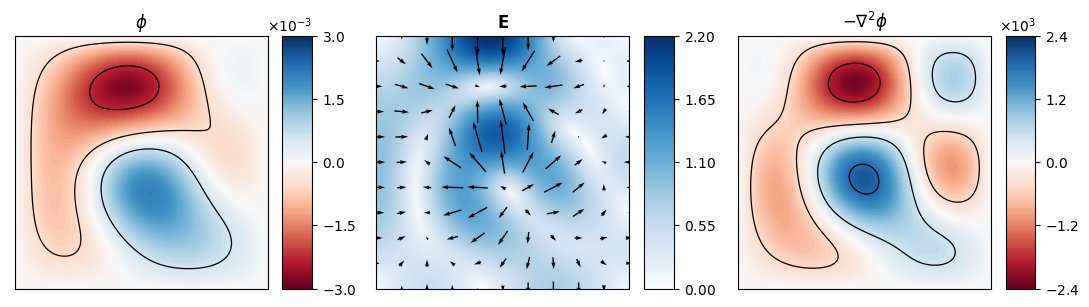}
    \caption{Example of Fourier dataset item with $N = M = 3$ modes and $p=0$ in a 1 cm$^2$ domain.}
    \label{fig:fourier_1_3_0}
\end{figure}

One example of a Fourier dataset is shown in Fig.~\ref{fig:fourier_1_3_0}. Note that in this case, the low number of modes of the dataset ($N=M=3$) allows a clear correlation between the potential and the charge distribution contrary to Fig.~\ref{fig:random_8_dataset}. This dataset allows therefore to understand the frequency response of the network by selecting particular frequencies. From now on, \texttt{fourier\_N\_p} will be used to denote $(N, p)$-Fourier datasets.

\section{Single frame inference of the potential and electric field}

\label{sec:nn_inference}

This section focuses on the training and prediction of the single Poisson solution for the potential and electric field, \textit{i.e.}, without coupling with the unsteady plasma Euler equations. The behavior of the networks when changing the architecture, loss function and dataset is evaluated. Training was typically performed over 300 epochs with $101 \times 101$-resolution datasets containing 10 000 snapshots (8000 for training and 2000 for validation) in a 1 cm$^2$ domain. All the networks are constructed with around 100 000 parameters, where the number of filters per layer is changed to approximately match this number when the number of branches is changed. All evaluations are performed at epoch 300. A summary of the parametric study is given in Tab.~\ref{tab:study_overview}. All computations were carried out using in-house Nvidia Tesla V100 SXM2 32 Gb GPUs.

\begin{table}[hbtp]
    \centering
    \begin{tabular}{| c | c |}
        \hline
        Architecture & UNet, MSNet \\
        \hline
        Number of branches & 3, 4, 5 \\
        \hline
        Receptive field & 50, 75, 100, 150, 200 \\
        \hline
        Number of parameters & 100 000 \\
        \hline
        Training snapshots & 10 000 \\
        \hline
        Training resolution & $101 \times 101$ \\
        \hline
    \end{tabular}
    \caption{Overview of the parametric study}
    \label{tab:study_overview}
\end{table}

\subsection{Metrics}

\begin{figure}[htbp]
    \centering
    \includegraphics[width=\textwidth]{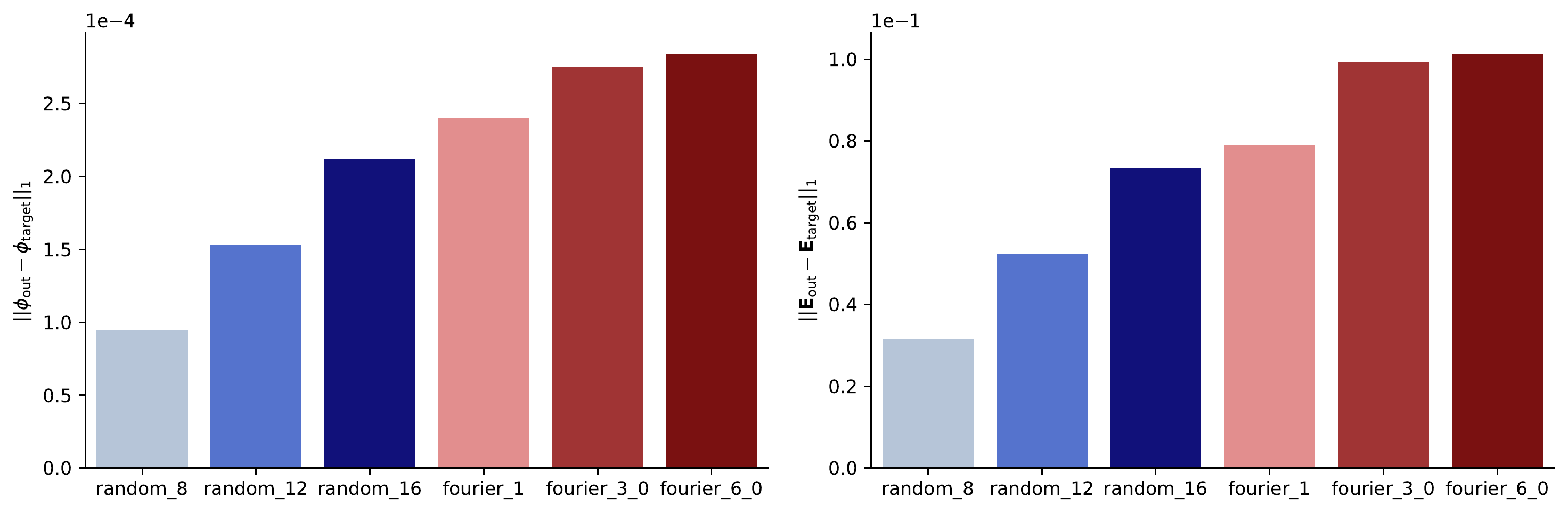}
    \caption{Datasets evaluation for 1-norm error of the potential (left) and electric field (right) with a UNet of $n_b = 3, k_s=3, \mrm{RF} = 100$.}
    \label{fig:oneitem}
\end{figure}

To monitor the accuracy of the networks the 1-norm and infinity norm residuals are used:

\begin{align}
    &\norm{u_\mrm{out} - u_\mrm{target}}_1 = \frac{1}{n} \sum_i \qty|u_\text{out}^i - u_\text{target}^i| \\
    &\norm{u_\mrm{out} - u_\mrm{target}}_\infty = \max_i \qty|u_\text{out}^i - u_\text{target}^i|
\end{align}

\noindent for $u \in \{\phi, \vb{E}\}$ and where the index $i$ spans all the relevant sizes of batch, dimension and directions x and y. The networks are evaluated on \texttt{random} and \texttt{fourier} datasets where each evaluation dataset contains 1000 snapshots. One network trained on \texttt{random\_8} snapshots and evaluated on the batch of datasets is shown in Fig.~\ref{fig:oneitem}. The accuracy of the network on Fourier datasets slightly decreases with an increasing number of modes $N$ and is also observed when $c$ increases for the random datasets. Overall both metrics give similar levels on all the datasets and are on the same order of magnitude not showing overfitting on the training dataset. From now on datasets are not expanded and a combined dataset evaluation is implied, \textit{i.e.}, the accuracy of the network is evaluated on the concatenation of the 6 datasets show in Fig.~\ref{fig:oneitem}.

Trainings have been carried out with both \texttt{random} and \texttt{fourier} datasets separately. \texttt{random}-trained networks are unambiguously better with both metrics and losses than \texttt{fourier}-trained networks. \texttt{random} datasets seem to contain more information about the Poisson equation and have therefore been preferred for training in all the following discussions while \texttt{fourier} datasets have been retained for \textit{a posteriori} analysis.

\subsection{Physical loss}

The \texttt{LaplacianLoss} depends only on the input and output of the network and does not need any target value. It is supplemented by the \texttt{DirichletLoss} to give a reference for the potential.

In a numerical simulation, the quantity of interest is the electric field $\vb{E}$ which as a derivative of the potential requires sufficient smoothness of the potential solution. Moreover, the electric field appears as a source term in the species $i$ momentum equation, which in turn impacts the species $i$ mass density (first term of $\vb{F}_i$). As a consequence, second-order smoothness on the potential must be ensured for the simulation to be stable.

\begin{figure}[htbp]
    \centering
    \includegraphics[width=0.8\textwidth]{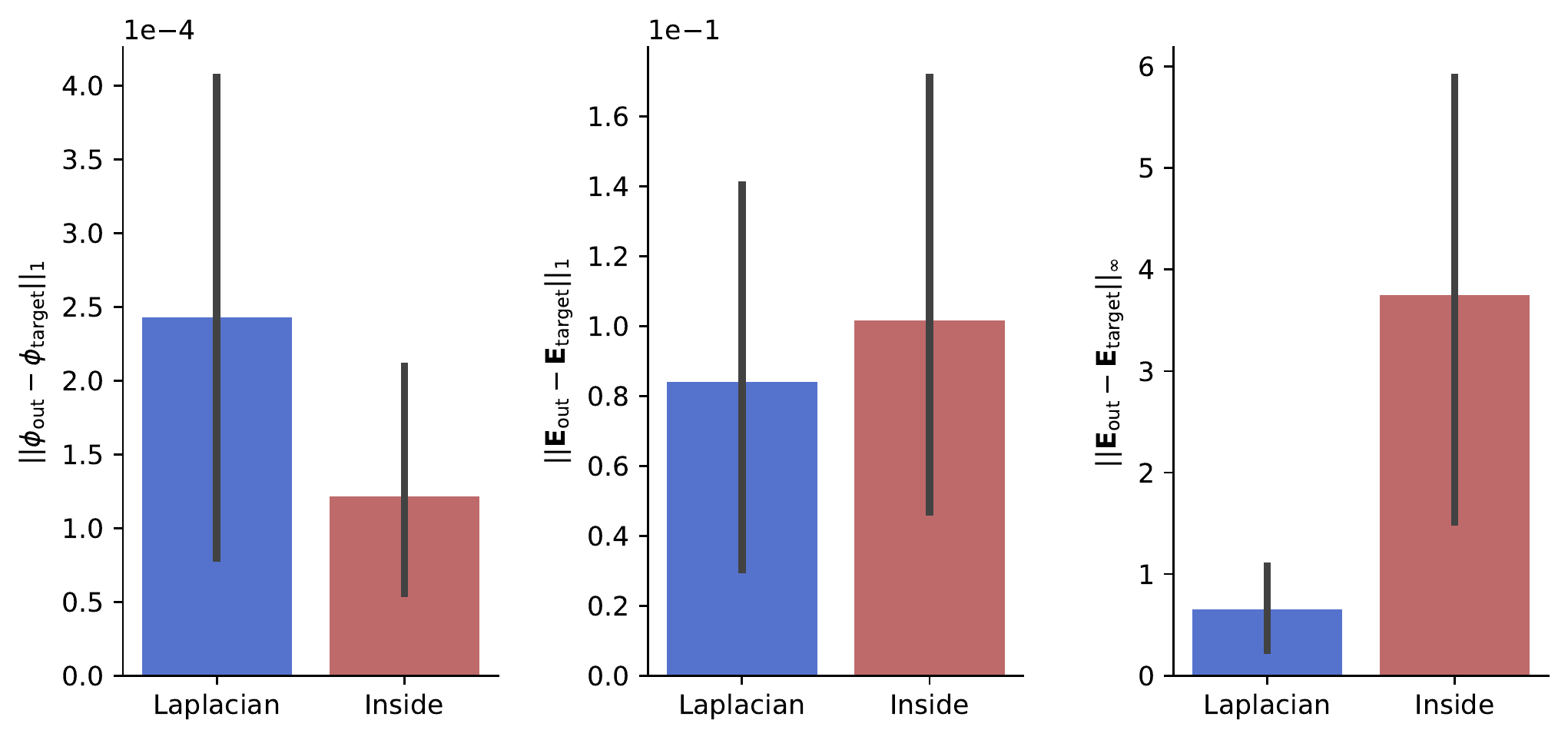}
    \caption{Comparison of \texttt{InsideLoss} and \texttt{LaplacianLoss} across \texttt{random\_8}-trained UNets with number of branches from 3 to 5 and RFs from 50 to 200. 1-norm of the potential (left), 1-norm (center) and infinity-norm (right) of the electric field.}
    \label{fig:physical_loss}
\end{figure}

Comparing the application of the \texttt{InsideLoss} and \texttt{LaplacianLoss} in training in Fig.~\ref{fig:physical_loss} shows a better accuracy on the 1-norm and infinity-norm (not shown here but similar) on the potential with \texttt{InsideLoss}. On the electric field 1-norm, both losses are comparable with slightly better accuracy for the \texttt{LaplacianLoss}. However the infinity-norm of $\vb{E}$ obtained with the \texttt{InsideLoss} is one order of magnitude greater than with the \texttt{LaplacianLoss}. This indicates few points where the electric field is unphysical and cannot be tolerated in the numerical simulation of the Euler plasma equations.

\begin{figure}[htbp]
    \begin{subfigure}[b]{\textwidth}
        \includegraphics[width = \textwidth]{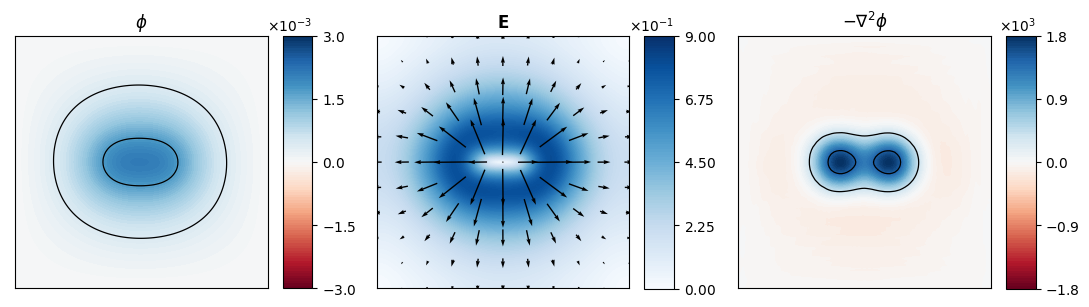}
        \caption{\texttt{LaplacianLoss}}
        \label{fig:laplloss_example}
    \end{subfigure}
    \begin{subfigure}[b]{\textwidth}
        \includegraphics[width = \textwidth]{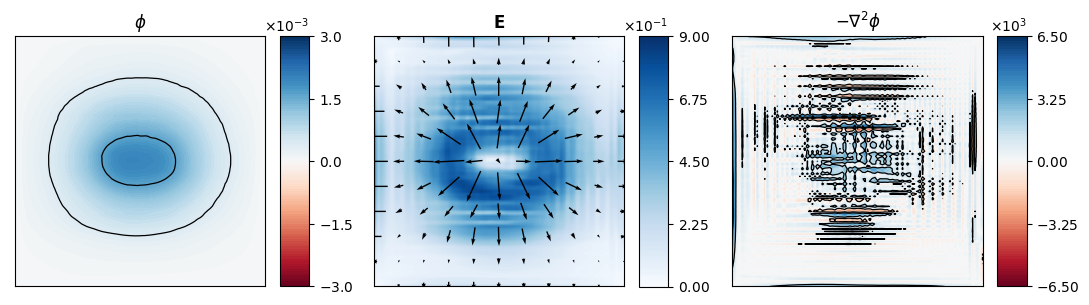}
        \caption{\texttt{InsideLoss}}
        \label{fig:insideloss_example}
    \end{subfigure}
    \caption{Potential [V] and electric field [V.m$^{-1}$] (norm in color and direction with arrows) associated to a two-Gaussians charge density [V.m$^{-2}$] in a 1 cm$^2$ domain for UNet with $n_b=3$, $\mrm{RF} = 100$. Top: \texttt{LaplacianLoss}. Bottom: \texttt{InsideLoss}.}
    \label{fig:laplvsinside}
\end{figure}

The inference of the same network, either trained with \texttt{LaplacianLoss} or with \texttt{InsideLoss} highlights the smoothing effect of the \texttt{LaplacianLoss}, while the Laplacian of the infered potential for an \texttt{InsideLoss}-trained network is completely unphysical. Comparing Fig.~\ref{fig:laplvsinside}(top) with Fig.~\ref{fig:problem_example} there is a very good agreement between the \texttt{LaplacianLoss} and linear solver solution. Consequently \texttt{LaplacianLoss} with \texttt{DirichletLoss} are chosen from now for all subsequent cases.

\subsection{UNet vs MSNet}

\begin{figure}[htbp]
    \centering
    \includegraphics[width=0.8\textwidth]{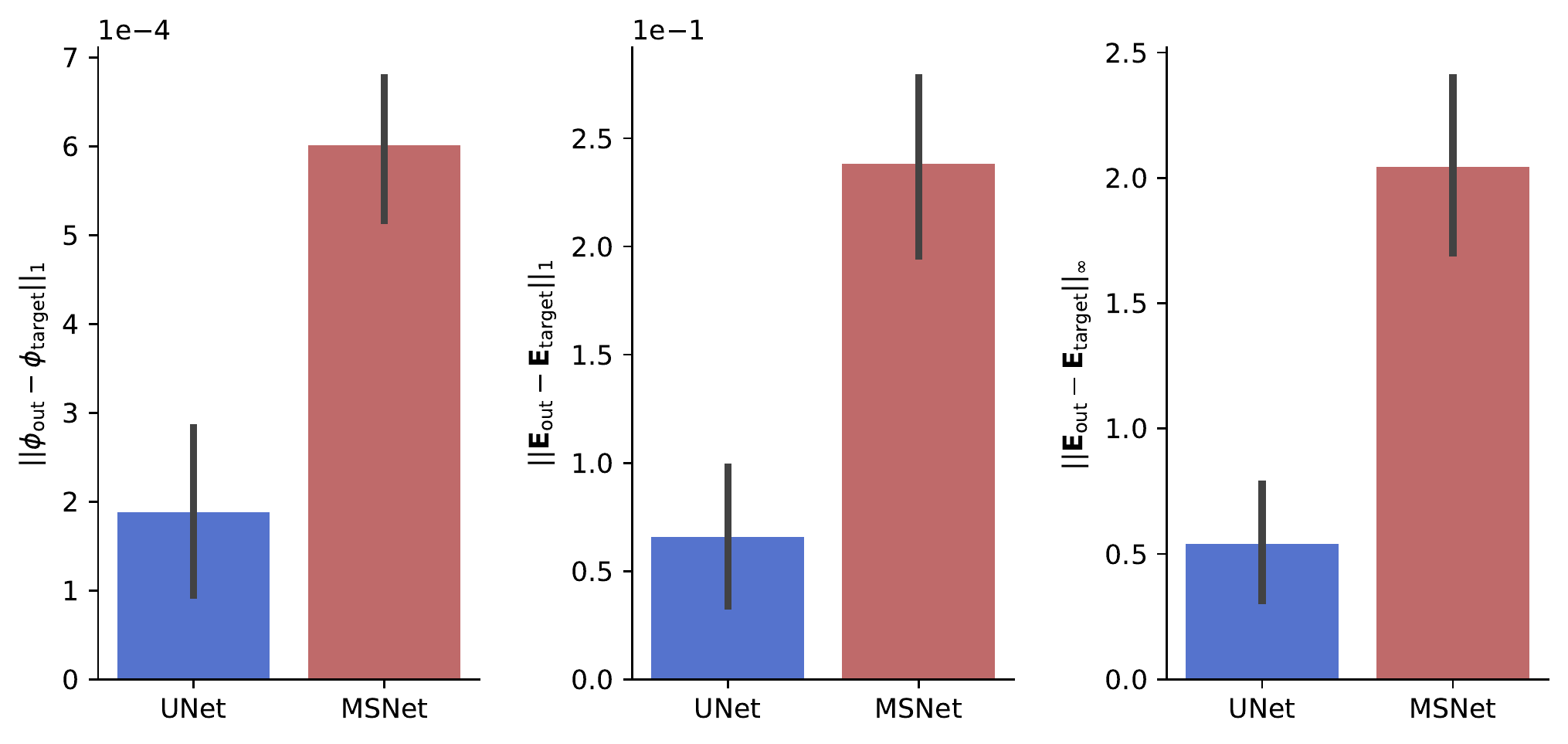}
    \caption{Comparison of MSNet and UNet architectures for different receptive fields [50-200] and numbers of branches [3-5]. 1-norm of the potential (left), 1-norm (center) and infinity-norm (right) of the electric field.}
    \label{fig:unetvsmsnet}
\end{figure}

UNet and MSNet architectures are compared in this section. The same number of parameters (around 100 000) and the same receptive fields and numbers of branches are used for both architectures. Comparison is shown in Fig.~\ref{fig:unetvsmsnet}. From these results, it appears that UNet architectures are better suited than MSNet architectures for the problem at hand over all parameters used. Thus only UNet architecture will be considered in the following.

One way to explain this difference may be found in the way each architecture goes from one scale to the other. MSNet has been designed to make video prediction \cite{multi_scale_paper} from one frame to the other and understand local movement, so only local information propagates from one snapshot to the other. Looking at Fig.~\ref{fig:msnet_sketch}, MSNets compact the information of one scale in one feature map before inserting it to the next scale. On the other hand, UNets in Fig.~\ref{fig:unet_sketch} apply a skip connection as well as an upsampling at the end of every intermediate scale, always keeping all relevant information.

\subsection{Receptive field}

\label{sec:receptive_field}

\begin{figure}[htbp]
    \centering
    \includegraphics[width=\textwidth]{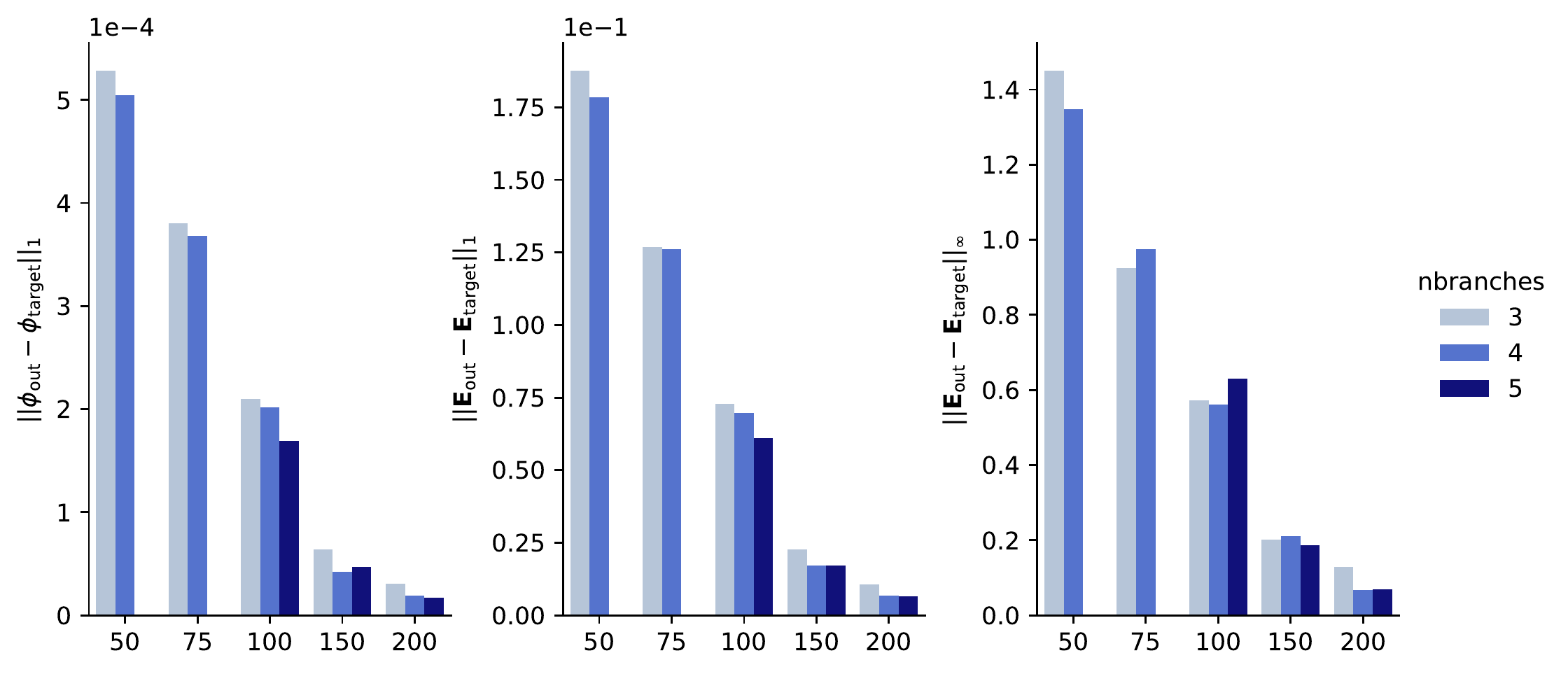}
    \caption{Error metrics of UNet for different receptive fields and numbers of branches from 3 to 5 branches and receptive fields from 50 to 200. 1-norm for the potential (left), 1-norm (center) and infinity-norm (right) for the electric field.}
    \label{fig:classic-rf50-200}
\end{figure}

Figure~\ref{fig:classic-rf50-200} shows the evaluation of UNet on the same resolution as the trained resolution, containing 3 to 5 branches, where the receptive fields vary from 50 to 200. Due to the definition of the receptive field, $\mrm{RF} = 100$ corresponds to the situation where the middle point of the input images can influence the whole output solution. The boundary pixels, however, do not influence the whole domain yet but only one quarter. Only when the receptive field reaches 200, any point of the input image influences the whole output domain, including the boundaries.

A first look at the results shows that networks with the same receptive have a similar behavior whatever the number of branches and depth, which therefore do not influence by themselves the network performance. This can be understood by looking at the structure of the Poisson equation: elliptic differential equation solutions need the information of the whole domain at every point. This elliptic nature is highlighted in the analytical solution of the problem where domain integrals are present to compute the Fourier coefficients of the charge density $R_{nm}$ Eq.~\eqref{eq:R_coefficients}.

Moreover, due to the potential damping Eq.~\eqref{eq:pot_coefficients} in $n^2$ and $m^2$, the low frequencies have the highest amplitudes. So it is critical for the network to be able to capture the whole domain when going through convolutions hence the importance of the receptive field. Thus accuracy improves when the receptive field increases because the network is able to capture the dominant longer wavelength content.

Accuracy is similar for different number of branches when keeping the receptive field constant. There is however a performance gain in higher number of branches networks. The convolutions are applied on lower resolution images, decreasing substantially the inference time at fixed number of parameters.

\begin{figure}[htbp]
    \centering
    \includegraphics[width=\textwidth]{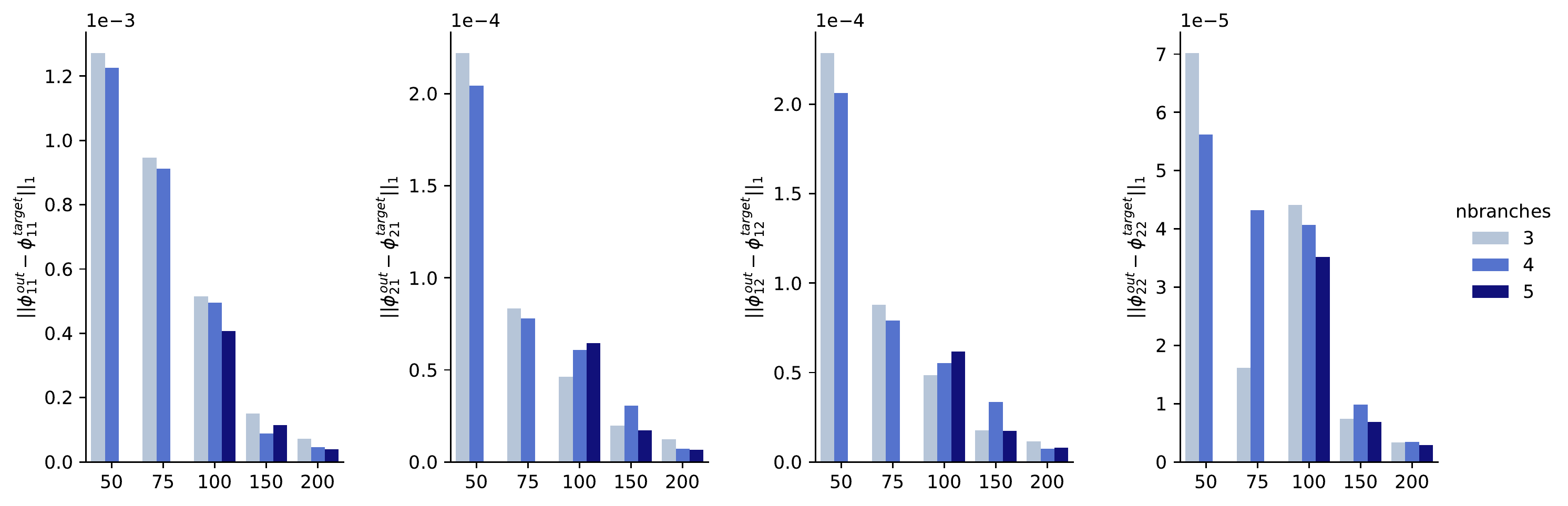}
    \caption{Error of amplitude of the first two modes of UNet solution for different receptive fields and numbers of branches.}
    \label{fig:fourier-rf50-200}
\end{figure}

A Fourier decomposition of the network outputs and target potential field is carried out to confirm this interpretation. The amplitudes of the first two modes are shown in Fig.~\ref{fig:fourier-rf50-200}. The amplitude of the fundamental mode $\phi_{11}$ follows the same trend as the other metrics and is at least an order of magnitude greater than the other first modes $\phi_{12}$, $\phi_{21}$, $\phi_{22}$ indicating that it drives the errors. Having a high receptive field should mostly impact all modes with $n=1$ or $m=1$, \textit{i.e.}, modes that contain a wavelength equal to the length of the domain.

\begin{figure}[htbp]
    \centering
    \includegraphics[width=\textwidth]{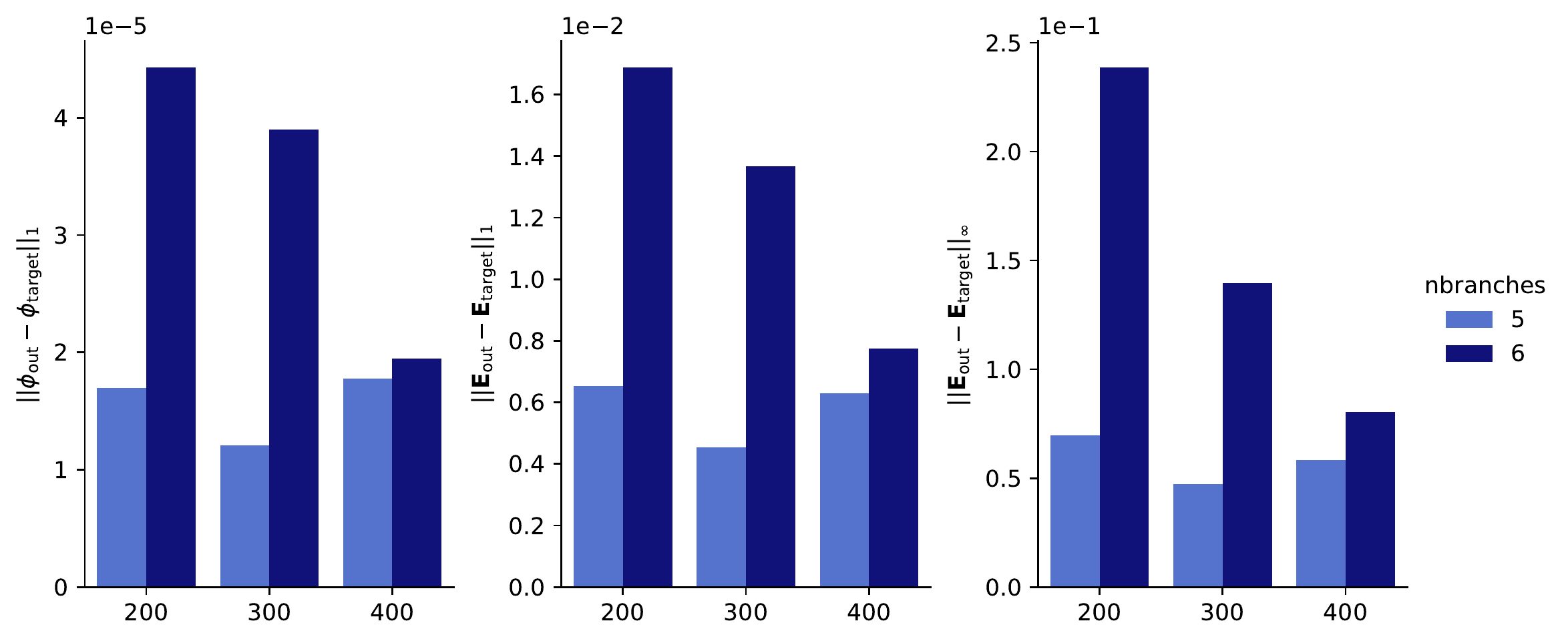}
    \caption{Error metrics of UNet for different receptive fields with high number of branches.}
    \label{fig:unet5-6}
\end{figure}

As explained in the beginning of the subsection, the monotonic decrease of the residuals from $\mrm{RF} = 50$ to $\mrm{RF} = 200$ is due to saturation of the domain of influence of every point. No significant increase of accuracy should be observed for higher receptive fields as the fundamental mode would not be better captured. This is shown in Fig.~\ref{fig:unet5-6} for UNet5 from $\mrm{RF} = 200$ to $\mrm{RF} = 400$.

As discussed previously networks with higher number of branches have faster resolution times due to faster convolutions when the number of points is lowered in the downscaled branches. There is however a limit to the number of branches due to the lowest resolution branch. The size of the lowest resolution branch images ($n_p/2^{n_b-1}$) should be greater than the kernel size ($k_s$) of the convolutional layers of that branch otherwise no information propagation is taking place (i.e. the most downscaled input image is smaller than the kernel size). For the $101 \times 101$ resolution images this means that UNet5 is an optimum because $\lfloor 101 / 2^4 \rfloor = 6 > k_s = 3$ whereas for UNet6 $\lfloor 101 / 2^5 \rfloor = 3 = k_s$. Thus adding a new branch at fixed number of network parameters is detrimental for the accuracy of the network. In the case of the UNet6, weights and biases used in the last branch ($b = 5$) are useless because no relevant information can be extracted from this scale. This is demonstrated in Fig.~\ref{fig:unet5-6} where the accuracy of the UNet6 is significantly less than UNet5 at the same receptive field. A gain of accuracy is observed for UNet6 when the receptive field increases whereas it is constant for UNet5. Since the weights of branch $b = 5$ are meaningless, the UNet6 acts as a reduced UNet5 with less parameters and a lower receptive field. The contribution to the receptive field of branch 5 $\mrm{RF}_5$ of the three UNet6-$\mrm{RF}200/300/400$ are respectively: 64, 128 and 192. The resulting effective receptive fields $\mrm{RF}_\mrm{eff} = \mrm{RF} - \mrm{RF}_5$ are thus 136, 172 and 208 and the increased accuracy can be explained following the same reasoning as the beginning of the section.

To sum up, for maximizing accuracy the receptive field should be chosen to saturate the domain of influence of any input point. That way the network correctly captures the low spatial frequencies which are dominant in most of the real-engineering physical fields. For performance optimization, the number of branches should be maximized as long as there is meaningful information in the downsampled branches (domain size bigger than the kernel size). Hence the optimal global parameters of the network for a given number of pixels $n_p$ should be:

\begin{align}
    \mrm{RF} &= 2 n_p \label{eq:optimal_rf}\\
    n_b &= \max \{ b \in \mathbb{N} | \lfloor n_p / 2^b \rfloor > k_s \} + 1 \label{eq:optimal_n_b}
\end{align}

\subsection{Resolution invariance and spectral analysis}
\label{sec:resolution_nn}

Thanks to the resolution scaling in Section \ref{subsec:res_scaling}, the network is able to work on resolutions different from the training resolution. Spectral analysis on different resolutions from a network trained on a single resolution is conducted in this section, where the optimum UNet5 with $\mrm{RF} = 200$ trained on $101 \times 101$ resolution images is used.

\begin{figure}[htbp]
    \centering
    \includegraphics[width=0.8\textwidth]{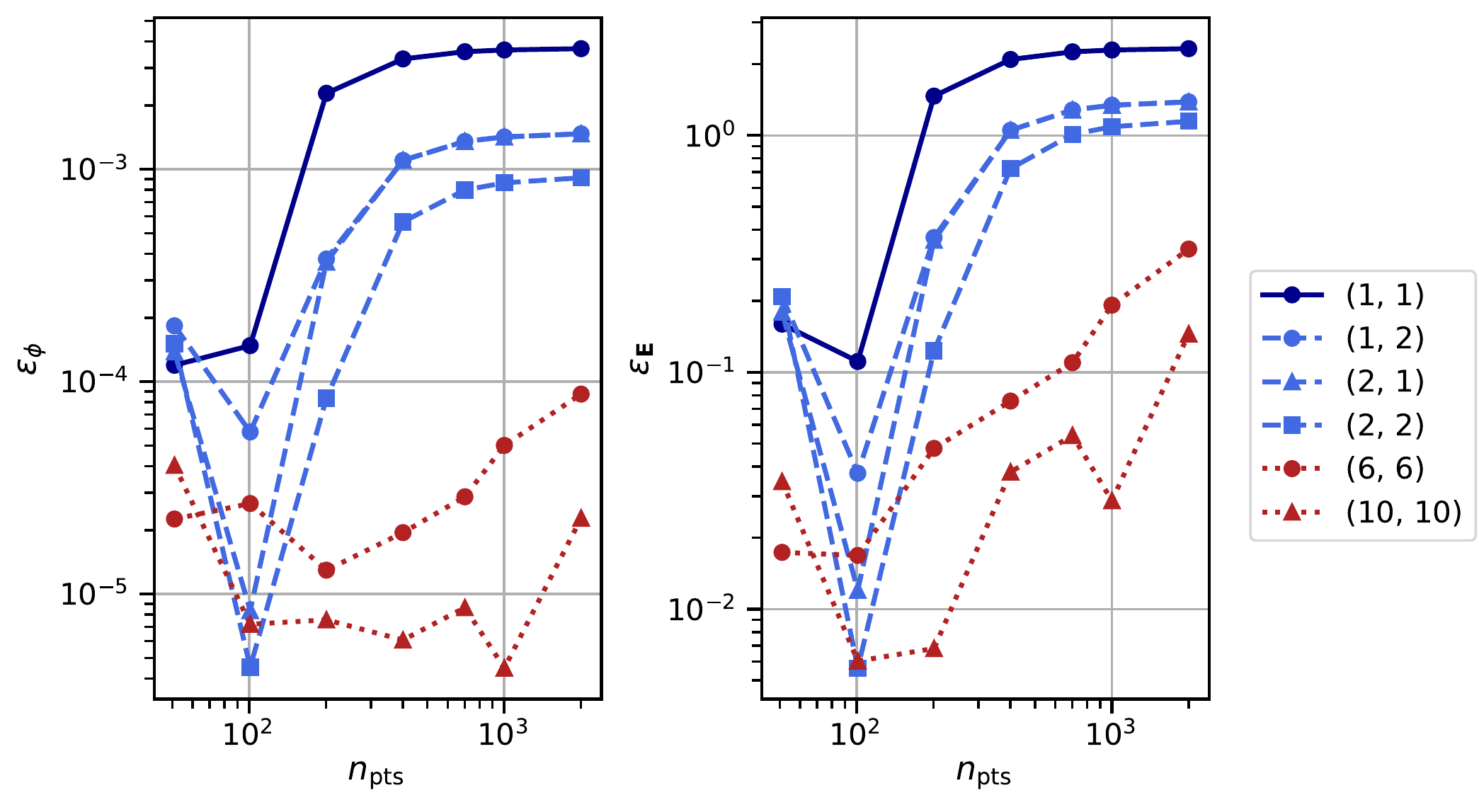}
    \caption{1-norm of the potential (left) and electric field (right) residuals for different modes $(n, m)$ and different resolutions with UNet5, $\mrm{RF} = 200$.}
    \label{fig:1norm-resolution-modes}
\end{figure}

The sine modes Eq.~\eqref{eq:natural_modes} for which the exact solutions are known are used to conduct the spectral analysis. Each mode is studied separately so that only one term $A_{nm}$ of the double sum in Eq.~\eqref{eq:natural_modes} is taken. The 1-norm residuals of the neural network potential and electric field of the UNet5-$\mrm{RF}$200 are shown in Fig.~\ref{fig:1norm-resolution-modes} for different values of $(n, m)$ as functions of the domain resolution.

The minimum residual is found at the trained resolution of 101 for both metrics. At that resolution, the $(n,m) = (1, 1)$ mode error is for the potential and electric field more than one order of magnitude higher than the $(n,m) = (10, 10)$ mode error. This tendency remains true at other resolutions so that the longer the wavelength the harder it is for the network to capture it correctly. Moreover, the frequency response of the network when increasing the resolution is not the same: the loss of accuracy for shorter wavelengths (red curves) is lower than higher wavelengths (blue curves).

The optimal network can thus work on resolutions that are different from the trained resolution, but errors grow as the tested resolution differs from the training one. One needs to keep in mind the loss of accuracy which is higher for longer wavelengths than shorter wavelengths. This loss of accuracy for resolutions that differ from the training resolution could be compensated either by interpolating the inference domain to the training resolution, by introducing a hybrid strategy~\cite{ajuria2021} which combines the network prediction with traditional iterative solvers to ensure a user-defined accuracy level or by training the network on multiple resolutions. However, these methods count with their own drawbacks that limit their use. On the one hand, interpolating the domain to the training resolution is a computationally expensive process, which considerably increments the time taken to complete the simulation. Moreover, even if the network accuracy increases on the interpolated domain, the interpolation introduces high frequency oscilations which are amplified by the network resulting in more unstable simulations. On the other hand, a hybrid strategy is suited to mitigate high frequency errors, as the charge field is locally difussed. However, iterative Jacobi solvers are not suited to cope with errors associated with long wavelengths, especially on high resolution domains. To correct the error related to low frequency modes, the number of needed Jacobi solver iterations is too high, considerably increasing the simulation time.

Note that finding strategies to make CNN work on multiple resolutions is still an open topic, which requires effort \cite{ozbay2021poisson,li2021fouriernet} and is out of the scope of the present paper. However, the guidelines obtained in this study, highlighting the key role of the receptive field and number of downscaled branches, are a first attempt to better understand how the CNN architecture learns the spatial distribution of the outputs. These guidelines could be reused to build efective CNN methods able to generalize on variable resolutions.

\subsection{Neural network performance}

Neural networks run best on GPUs whereas linear system solvers have been historically run on CPUs. To assess the neural network performance against classical linear solvers, CPU and GPU performances need to be compared. The methodology applied here is as follows: given a computational node containing CPUs and GPUs, the speedup when activating or not the GPUs compared to the use of all the CPUs available in one node is assessed. This indicates the potential speedup that the neural network can provide compared to a classical linear system solver running on the same CPUs. Code to run all the benchmarking presented in the following is available at \url{https://gitlab.com/cerfacs/plasmanet}.

Two configurations have been used in our local cluster: \texttt{config\_1} is a bi-socket Intel node with 2 x 18 core Xeon Gold 6140 (2.3 Ghz clock speed and 96 Gb memory) interacting with 4 NVIDIA V100 32 Gb GPUs where only one of the four GPUs is used in this study. The second configuration \texttt{config\_2} is a bi-socket AMD node with 2 x 64 core EPYC Rome 7702 (2 Ghz clock speed and 512 Gb of memory) interacting with a single NVIDIA A100 40 Gb GPU.

Therefore following the methodology, we use all the cores available in one computational node (36 for \texttt{config\_1} and 128 for \texttt{config\_2}) to assess the minimal resolution time of linear Poisson solvers using PETSc \cite{petsc-web-page}.

\begin{figure}[htbp]
    \centering
    \includegraphics[width=0.5\textwidth]{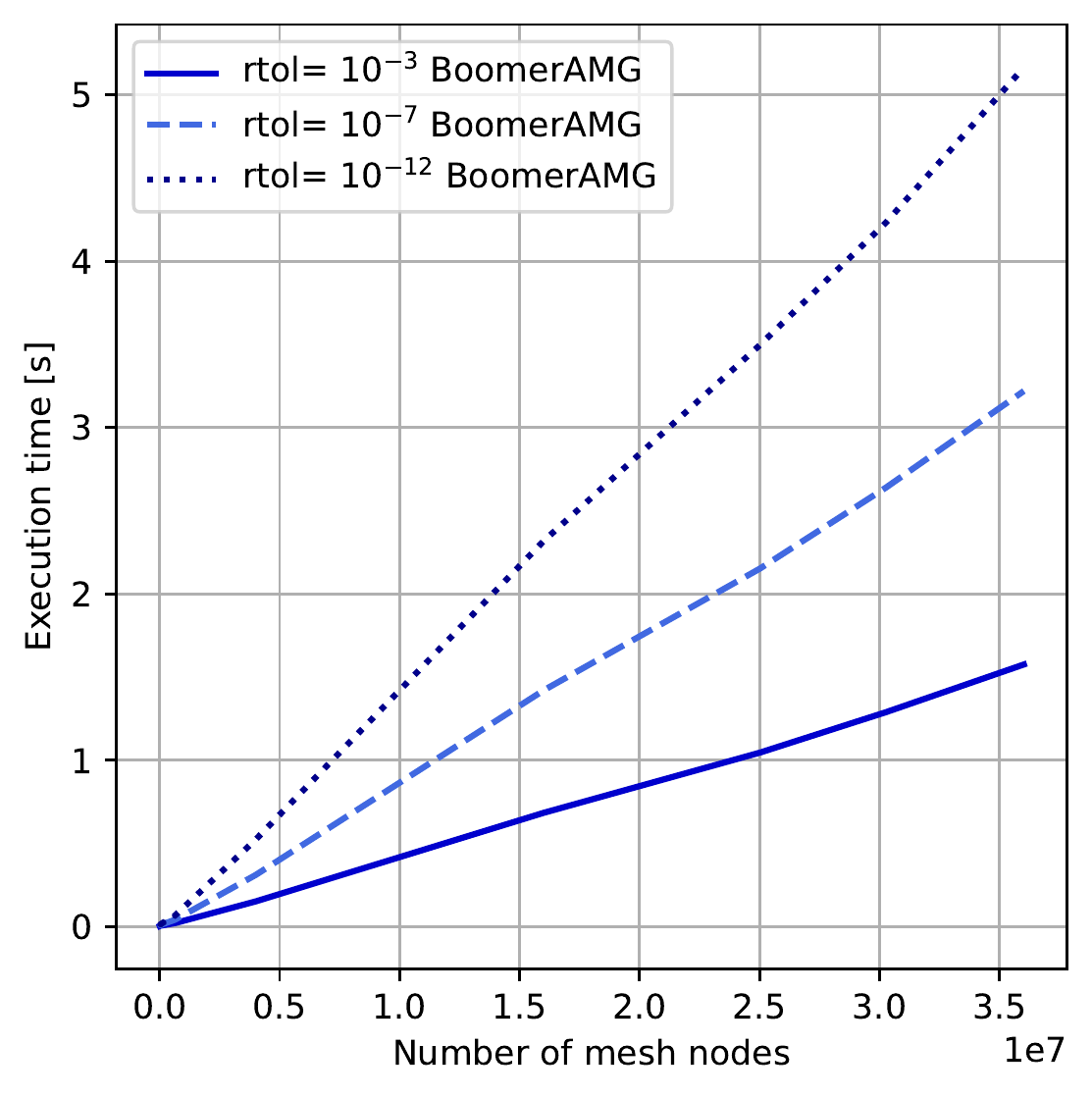}
    \caption{Performance of linear solver when raising the relative tolerance for different AMG solvers on \texttt{config\_1}.}
    \label{fig:rtol_perfs}
\end{figure}

The Dirichlet boundary conditions Poisson problem on the 2D square of \SI{1}{\centi\metre\squared} with two Gaussian charge density is used to compare the linear system solver and neural network solver performances. In the case of the linear system solver, the matrix has been symmetrized so that it is positive symmetric definite. Various linear solvers have been tested and results are presented in \ref{appendix:petsc_benchmark}. The Conjugate Gradient (CG) method \cite[Chap. 6.7]{saad_2003} as iterative solver and HYPRE BoomerAMG \cite{hypre_ref} preconditioner is the highest performing option in this case. To get closer to the accuracy of the neural network solver, the relative tolerance of the iterative solvers has been raised to $10^{-3}$ where a 4 times speed up is observed compared to a $10^{-12}$ relative tolerance as shown in Fig.~\ref{fig:rtol_perfs} for \texttt{config\_1}. All the execution times shown with PETSc or the neural network solver are averages taken over 20 resolutions.

\begin{figure}[htbp]
    \centering
    \begin{subfigure}[b]{0.45\textwidth}
        \centering
        \includegraphics[width=\textwidth]{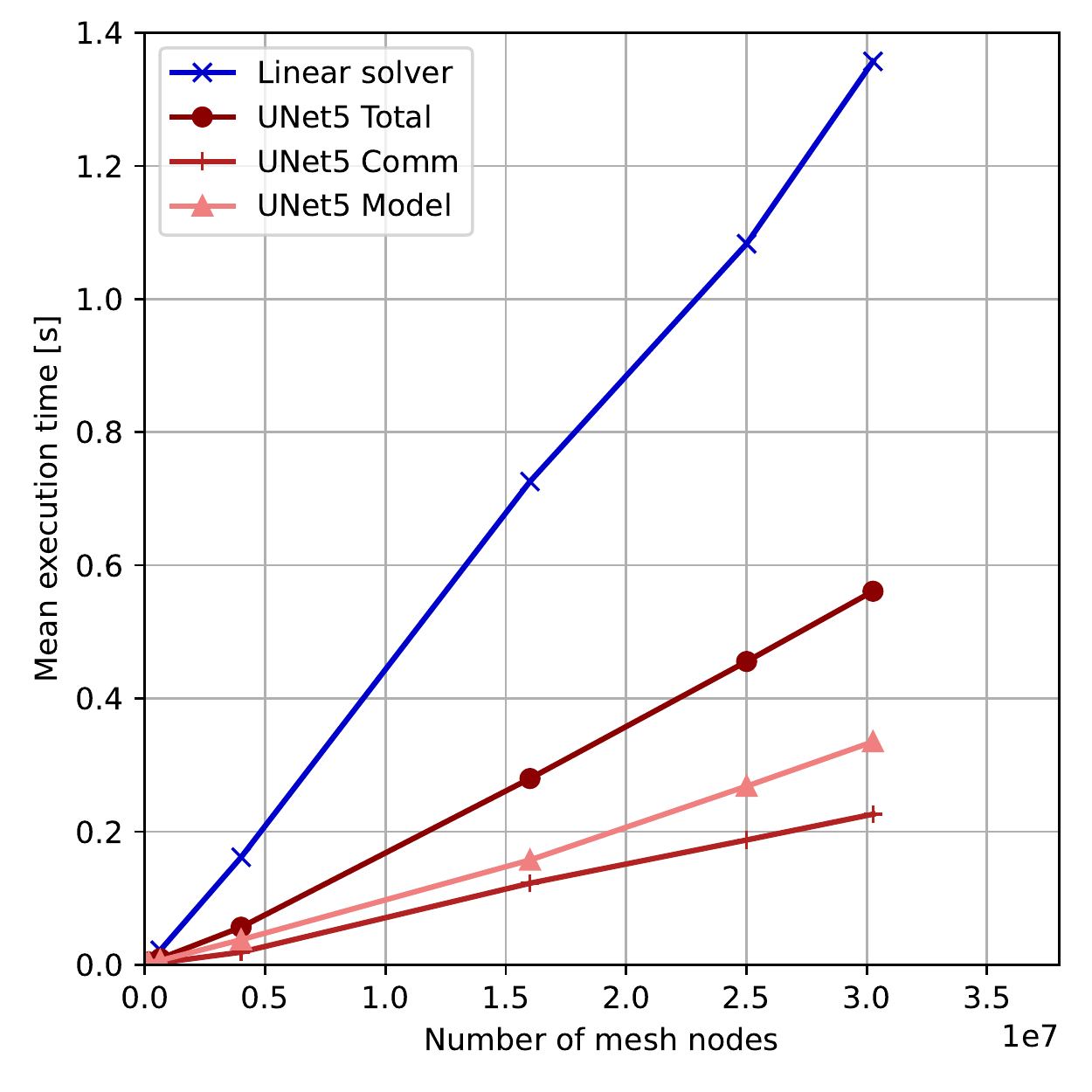}
        \caption{\texttt{config\_1}}
        \label{fig:V100}
    \end{subfigure}
    \begin{subfigure}[b]{0.45\textwidth}
        \centering
        \includegraphics[width=\textwidth]{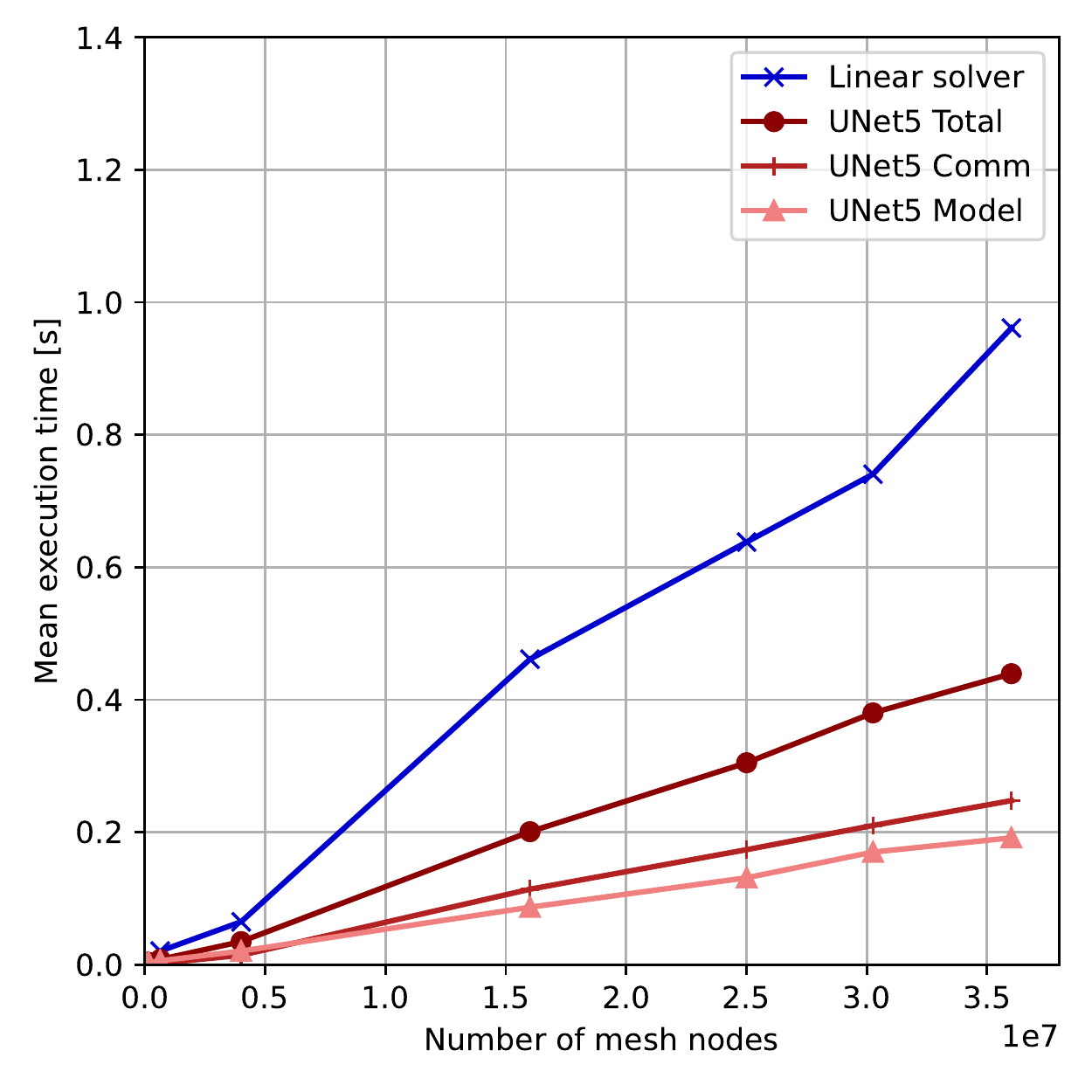}
        \caption{\texttt{config\_2}}
        \label{fig:A100}
    \end{subfigure}
    \caption{Performance of linear solver (CG-BoomerAMG) against neural network solver.}
    \label{fig:ls_vs_nn_perfs}
\end{figure}

The neural network performance compared to PETSc linear system solver using CG-BoomerAMG is shown in Fig.~\ref{fig:V100} on \texttt{config\_1} where the total execution times, model inference times and communication times of the network are shown. At high number of mesh nodes, the linear solver run time on 36 cores is higher than the neural network total run time by a factor of 2.5 for $30 \times 10^6$ mesh nodes. Concerning the GPU performance, the communication time increases with the number of nodes and becomes a significant part of the execution time, as expected. Note that only one of the four GPUs available on the computational node has been used as inference on multi-GPUs is not implemented for the neural network. The maximum resolution of $5501 \times 5501$ corresponds to the maximum memory of the GPU at hand (32 Gb) and depends on the hardware available. This is a clear limitation of the neural network solver as it is much more memory consuming than the classical linear system solver: the UNet5 network architecture used at $5501 \times 5501$ resolution use up around 30 Gb whereas a single float 64 array of $5501 \times 5501$ is around 200 Mb.

Results for \texttt{config\_2} are shown in Fig.~\ref{fig:A100}. The GPU memory is higher (40 Gb against 32 Gb for V100 GPU) and the number of cores available as well (128 against 36 for \texttt{config\_1}). The higher memory allowed the inference of a finer resolution at $6001 \times 6001$ (the point at $3.6 \times 10^7$ number of nodes in Fig.~\ref{fig:A100}). For this configuration, the resolution time of the neural network running on the A100 GPU is about a factor 2 lower than the linear system solver on 128 cores, making it a viable option in terms of performance.

Comparing the V100 and A100 GPUs, a similar communication time, \textit{i.e.} the time taken by the CPUs to send the data to the GPU, is observed. However the model time, \textit{i.e.} the application of the neural network on the GPU, is about two times faster for the A100 GPU compared to the V100 GPU.

Thus, deep neural networks are shown to effectively accelerate simulations. This study paves the way for future ones to further analyze and improve inference times and memory cost of the neural networks.

\section{Neural networks and canonical plasma fluid simulation}

\label{sec:nn_plasma_oscill}

From the previous section, at a resolution of $n_p=101$ pixels in each direction, a 100 000 parameter UNet architecture trained on a \texttt{random\_8} dataset, with receptive field $\mathrm{RF}=5$ and number of branches $n_b=5$ using a combined \texttt{LaplacianLoss}-\texttt{DirichletLoss} is chosen for best network accuracy and performance. The target test case corresponds to the plasma oscillations in a square domain, in order to have analytical solutions of this space-time evolving plasma problem to assess both accuracy and performance of the proposed method. The network solves the Poisson equation in place of the linear system solver coupled to the plasma Euler equations. The performance of both solvers is fully analyzed.

\subsection{2D plasma oscillation test case}
\label{section:electron_plasma_oscillation}

One of the fundamental properties of plasmas is to maintain electric charge neutrality
at a macroscopic scale under equilibrium conditions. When this macroscopic charge neutrality is
disturbed, large Coulomb forces come into play and tend to restore the macroscopic charge neutrality \cite[Chap. 11.1]{bittencourt}.

Electrons and positive ions with charge $e$ are considered. Ion motion is neglected since its mass is way larger than that of the electrons. A very small electron density perturbation $n_e$ is initialized such that:

\begin{align}
    n_\mrm{electron}(\mathbf{r},t) &= n_0 + n_e(\mathbf{r},t) \\
    n_\mrm{ion}(\vr, t) &= n_0
\end{align}

\noindent where $n_0$ is a constant number density and $|n_e|\ll n_0$. Linearization of the momentum equation, combined with the mass equation and the Maxwell-Gauss equation \cite[Chap. 11.1]{bittencourt} yields:

\begin{align}
    \frac{\partial^2n_e}{\partial t^2}+\omega_p^2 n_e = 0   \qq{where}
    \omega_p = \sqrt{\frac{n_e e^2}{m_e\varepsilon_0}}
\end{align}

The electron density varies harmonically in time at the electron plasma frequency $f_p = \omega_p / 2\pi$, or oscillation period $T_p = 1 / f_p$ and it can be shown that the electric field does as well.

Note that the initial electron perturbation $n_e(x, y, t = 0)$ can be chosen arbitrarily, in the end electron density and electric field profiles vary harmonically at pulsation $\omega_p$:

\begin{align}
  n_e(x, y, t) &= n_e(x, y, t = 0) \, \cos(\omega_p t) \\
  \vb{E}(x, y, t) &= \vb{E}(x, y, t = 0) \, \cos(\omega_p t)
\end{align}

This plasma oscillation can be simulated by discretizing the 2D plasma Euler equations in a cell-vertex formulation with a classical Law-Wendroff scheme (second order in time and space) \cite[Chap. 4]{lamarquethesis}.

Taking a typical value for the background density \cite{celes2008}, $n_0 = 10^{16}\,\SI{}{\metre\cubed}$ is used. This value gives an oscillation period of $T_p = \SI{1.11}{\nano\second}$. A perturbation amplitude around $n_e = 10^{11}\,\SI{}{\metre\cubed}$ is used. This value is not critical and only needs to satisfy $n_e \ll n_0$. In the cases presented the electron density field is initially perturbed with a two-Gaussians shape function.

\subsection{Neural network Poisson equation solver in plasma oscillation simulation}

The use of the neural network to solve the Poisson equation coupled with the unsteady Euler equations to simulate the 2D electron plasma oscillation is carried out using the different architectures and losses presented in the previous section. As for the test cases of the previous section, optimum results are obtained with UNet5, $\mrm{RF} = 200$ when coupled with transport equations. The choice of \texttt{LaplacianLoss} over \texttt{InsideLoss} is critical to get a stable simulation as already shown in Fig.~\ref{fig:laplvsinside}. A high enough receptive field is also necessary to get an accurate solution.

\begin{figure}[htbp]
    \begin{subfigure}[b]{\textwidth}
        \includegraphics[width = \textwidth]{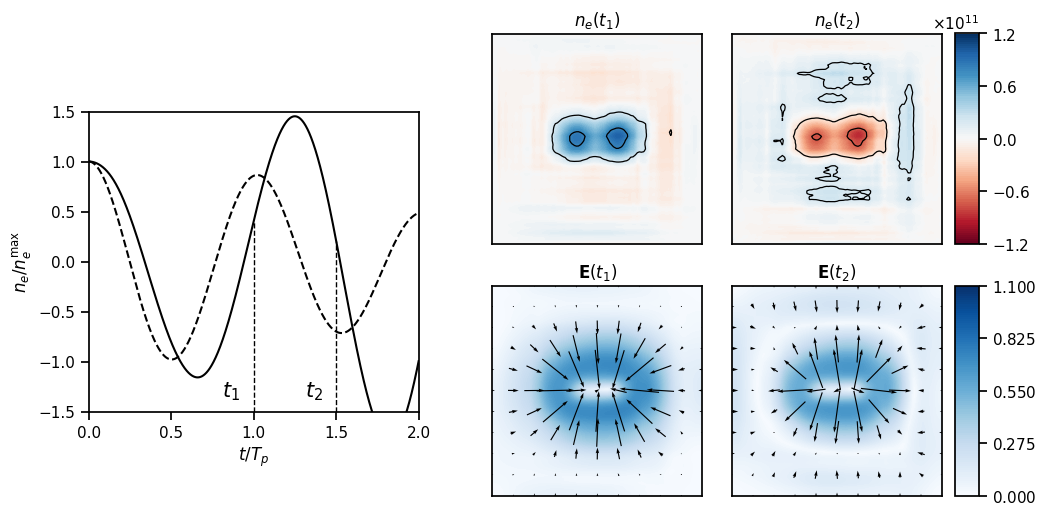}
        \caption{$\mrm{RF} = 100$}
        \label{fig:plasma_oscill_nn_rf100}
    \end{subfigure}
    \begin{subfigure}[b]{\textwidth}
        \includegraphics[width = \textwidth]{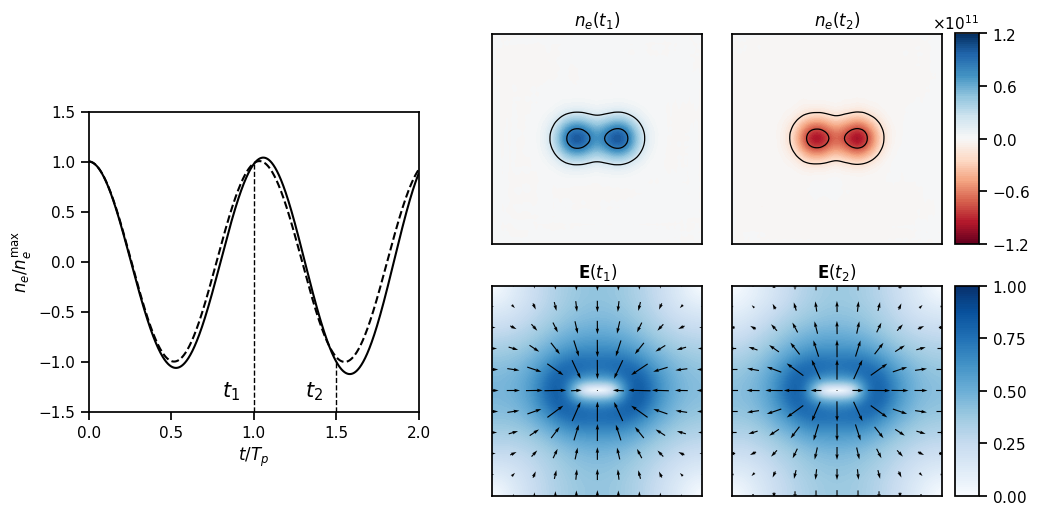}
        \caption{$\mrm{RF} = 200$}
        \label{fig:plasma_oscill_nn_rf200}
    \end{subfigure}
    \caption{Plasma oscillation simulation with the Poisson equation solved with UNet5 at different receptive fields. Temporal evolution of the mean (solid) and high absolute values (dashed - points inside the contours of electron density in Fig.~\ref{fig:problem_example}) on the left and snapshots of $n_e$ and $\vb{E}$ at $t_1 = T_p$ and $t_2 = 1.5 \, T_p$ on the right.}
    \label{fig:plasma_oscill_nn}
\end{figure}

Examples of plasma oscillation simulations where the Poisson equation has been solved by networks with different receptive fields are shown in Fig.~\ref{fig:plasma_oscill_nn}. Two UNet5 networks with  $n_b = 5$ and $\mrm{RF}=100,\, 200$ are used. As shown in Fig.~\ref{fig:classic-rf50-200}, a gain of accuracy with increased $\mrm{RF}$ is observed with a factor of around 4 for both 1 and infinity norms of the electric field from $\mrm{RF}=100$ to $\mrm{RF}=200$. This accuracy gain has a real impact on the solution as seen in Fig.~\ref{fig:plasma_oscill_nn_rf100} where the contours of electron density are not smooth anymore. On the other hand, the $\mrm{RF}=200$ network produces very satisfactory results in Fig.~\ref{fig:plasma_oscill_nn_rf200}. Finally the quantity of interest of the simulation which is the plasma oscillation period $T_p$, is not well captured in Fig.~\ref{fig:plasma_oscill_nn_rf100} whereas it is perfectly retrieved in Fig.~\ref{fig:plasma_oscill_nn_rf200}.

\section{Double headed streamer}

The previous test case of plasma oscillation, although representative of the interaction between electromagnetic field and plasma species, does not include any chemistry or numerical stiffness as the perturbation electron density at the origin of the plasma oscillation is five orders of magnitude smaller than the background density. A more complex and stiffer case is proposed here to validate the whole methodology developed in the previous sections: the double headed streamer introduced in \cite{celes2008}. Streamer discharges are relevant in plasma assisted combustion \cite{Ju2016} and material processing \cite{lieberman}.

Transport and chemistry kinetic coefficients are detailed in \cite{Morrow1997}. This chemistry was used to model atmospheric plasma discharges in air in Celestin \cite{celes2008} and Tholin \cite{tholin2012} among others. It is composed of three species: electrons ($n_e$), positive ions ($n_p$) and negative ions ($n_n$). Those three species are modelled in a drift-diffusion approximation so that only densities need to be monitored, which is a reasonably well approximation in plasma discharges \cite{celes2008}. The electrons are much faster than the ions due to the mass ratios, so that during the time of the discharge propagation, the ions ($n_p$ and $n_n$) can be considered not moving: no transport for them is required, and they are therefore only affected by chemistry, which depends on the magnitude of the electric field $E = |\vb{E}|$. Thus, the system of equations reads:

\begin{align}
\frac{\partial n_e}{\partial t} + \nabla \cdot \qty(n_e \mathbf{W_e} - D_e \nabla n_e) &= n_e \alpha |W_e| - n_e \eta |W_e| - n_e n_p \beta \\
\frac{\partial n_p}{\partial t} &= n_e \alpha |W_e|-n_e n_p \beta-n_n n_p \beta\\
\frac{\partial n_n}{\partial t} &=  n_e \eta |W_e| - n_n n_p \beta
\end{align}

\noindent where $\alpha = \alpha(E/N)$ is the ionization coefficient, $\eta = \eta(E/N)$ the attachment coefficient, $N$ the neutral gas density, $\beta$ the recombination rate, $\vb{W}_e = - \mu_e \vb{E}$ the drift-velocity of the electrons and $\mu_e = \mu_e(E/N)$ the electron mobility. The electric field $\vb{E}$ is critical as it controls both transport for electrons and chemistry for all species. Analogously to the plasma Euler equations, the electric field is computed from the potential given by the Poisson equation:

\begin{gather}
    \nabla^2 \phi = -\frac{\rho}{\epsilon} \qq{where} \rho = e(n_p - n_e - n_n) \\
    \vb{E} = -\nabla {\phi}
\end{gather}

The double headed streamer is initialized with a neutral Gaussian profile at $x = 2$ mm and $r = 0$ mm with a background density in a rectangular domain of $L_x \times L_r = 4 \times 1$ mm$^2$, corresponding to an azimuthal cut of the cylindrical geometry, so that

\begin{equation}
    n_e = n_p = n_0 \exp[-\qty(\frac{x - x_0}{\sigma_x})^2 - \qty(\frac{r}{\sigma_r})^2] + n_\mrm{back}
\end{equation}

\noindent with $n_0 = \SI{e19}{\per\cubic\meter}$, $n_\mrm{back} = \SI{e14}{\per\cubic\meter}$ and a strong constant electric field of $E_x = \SI{4.8e6}{\volt\per\meter}$ is applied at the boundary conditions.

A robust upwind scheme has been adopted for the advection part of the electron density with central differencing for the diffusion flux and Euler time integration is performed with a timestep of $\Delta t = 10^{-12}$ s.

The transport equations and the Poisson equation are solved in cylindrical coordinates. An axisymmetric formulation is used so that the 2D domain corresponds to a uniform grid of coordinates $(x, r)$. Because of the cylindrical coordinates, solving the Poisson problem is different compared with the previous 2D cartesian problem (Sections \ref{sec:nn_inference} and \ref{sec:nn_plasma_oscill}):

\begin{empheq}[left=\empheqlbrace]{align}
    \nabla^2 \phi &= \frac{1}{r} \pdv{r}\qty(\frac{1}{r}\pdv{\phi}{r}) + \pdv[2]{\phi}{x} = - R \qq{in} \mathring{\Omega} \\
      \phi &= - E_x x \qq{on} \partial \Omega_D \\
      \nabla \phi \cdot \vb{n} &= 0 \qq{on} \partial \Omega_N
\end{empheq}

\noindent where Dirichlet boundary conditions are applied at $x = 0$, $x = L_x$, $r = L_r$ and Neumann boundary conditions are applied at the axis $r = 0$. A loss function \texttt{NeumannLoss} has been introduced to take into account this new boundary condition:

\begin{equation}
    \label{eq:neumannloss}
    \mathcal{L}_N(\vb{\phi}_\text{out}) = \frac{1}{b_s (n_x - 2)} \sum_{b, i} (\nabla \phi_\text{out}^{b, 0, i} \cdot \vb{e}_r)^2
\end{equation}

\noindent so that three losses are used in this case: \texttt{NeumannLoss}, \texttt{DirichletLoss} and \texttt{LaplacianLoss}. Note that a constant background electric field $E_x$ is applied at the boundary conditions. To stay close to the previous study the problem has been split in two: the neural network deals with zero Dirichlet boundary conditions and charge density. The rest of the problem only yields a constant electric field $E_x \vb{e}_x$. From the superposition principle the total electric field is the sum of the neural network solution and the constant electric field:

\begin{equation}
    \vb{E} = \vb{E}_\mrm{NN} + E_x \vb{e}_x
\end{equation}

\begin{figure}[htbp]
    \centering
    \includegraphics[width=\textwidth]{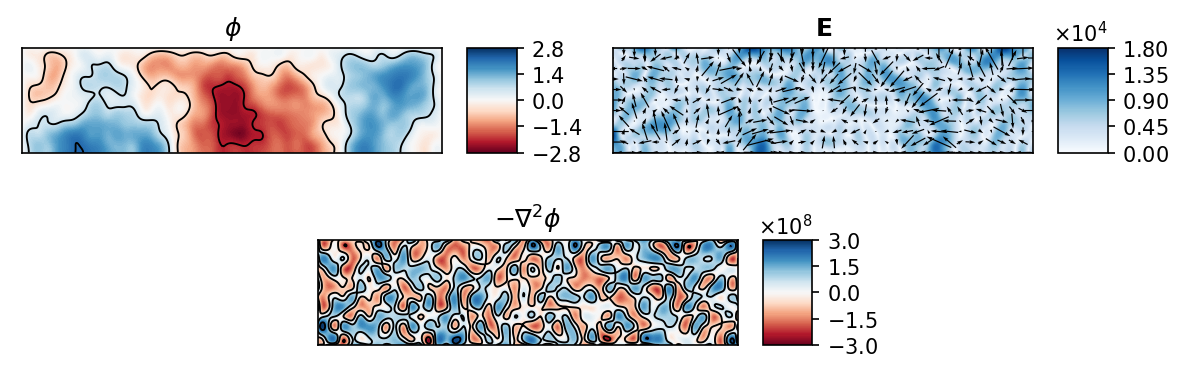}
    \caption{Example of \texttt{random\_8} source term input in a  $4 \times 1$ mm$^2$ cylindrical domain.}
    \label{fig:random_8_cyl_dataset}
\end{figure}

Training with random profiles as described previously has been done, where a sample of the dataset is shown in Fig.~\ref{fig:random_8_cyl_dataset}. Unlike the cartesian geometry case, the potential is not uniformly distributed but it is amplified at the axis $r=0$ corresponding to the bottom of the 2D domain in Fig.~\ref{fig:random_8_cyl_dataset}.

The network architecture has been chosen accordingly to the prescribed optimal parameters of Eqs.\eqref{eq:optimal_rf} and \eqref{eq:optimal_n_b}. These have been adapted to the present case as the geometry is now rectangular and not squared so that a receptive field in each direction can be defined. To achieve these guidelines on the $401 \times 101$ mesh, receptive fields of $\mrm{RF}_x = 800$ and $\mrm{RF}_y = 200$ have been chosen with $n_b = 5$ branches and around 100 000 parameters.

\begin{figure}
    \centering
    \begin{subfigure}[b]{0.45\textwidth}
        \centering
        \includegraphics[height=6.5cm]{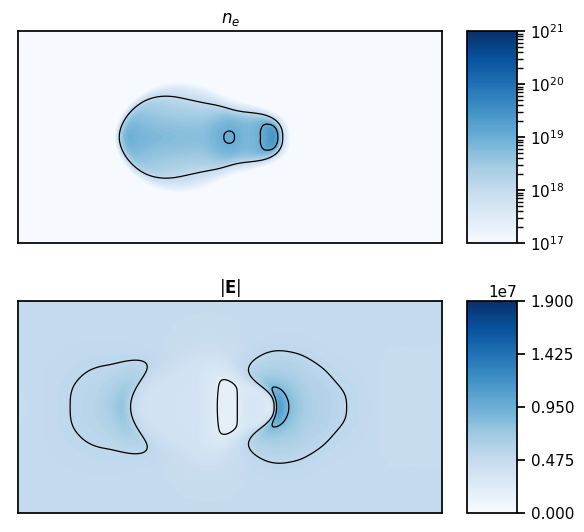}
        \caption{Neural network}
    \end{subfigure} \hspace{0.2cm}
    \begin{subfigure}[b]{0.45\textwidth}
        \centering
        \includegraphics[height=6.5cm]{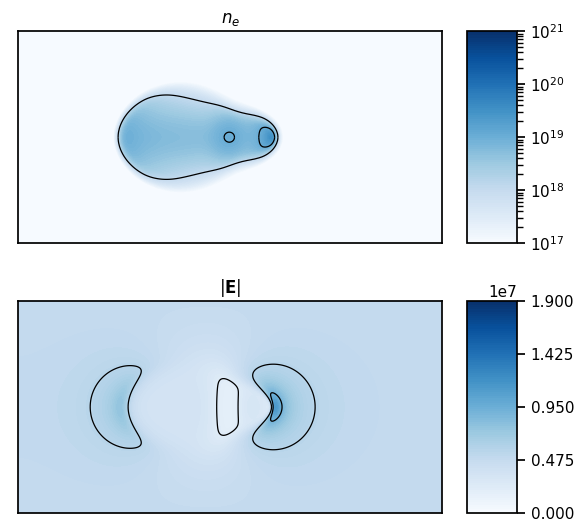}
        \caption{Linear system}
    \end{subfigure}
    \caption{Comparison of electron density and electric field norm at 1.6 ns for neural network and linear system Poisson solver. The computational domain has been mirrored from the central axis.}
    \label{fig:comparison_streamer_nn_vs_ls_16ns}
\end{figure}

\begin{figure}
    \centering
    \begin{subfigure}[b]{0.45\textwidth}
        \centering
        \includegraphics[height=6.5cm]{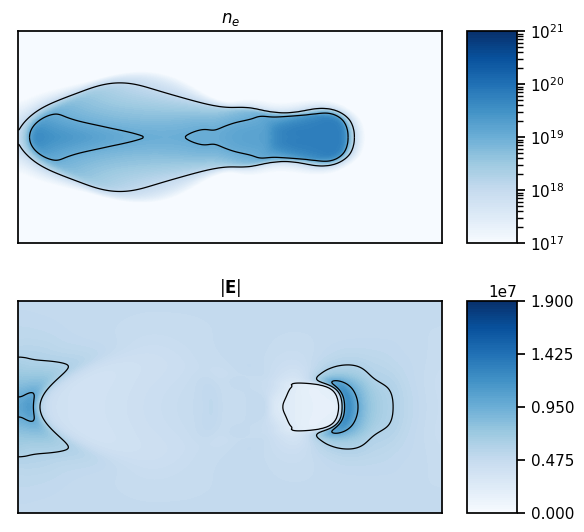}
        \caption{Neural network}
    \end{subfigure} \hspace{0.2cm}
    \begin{subfigure}[b]{0.45\textwidth}
        \centering
        \includegraphics[height=6.5cm]{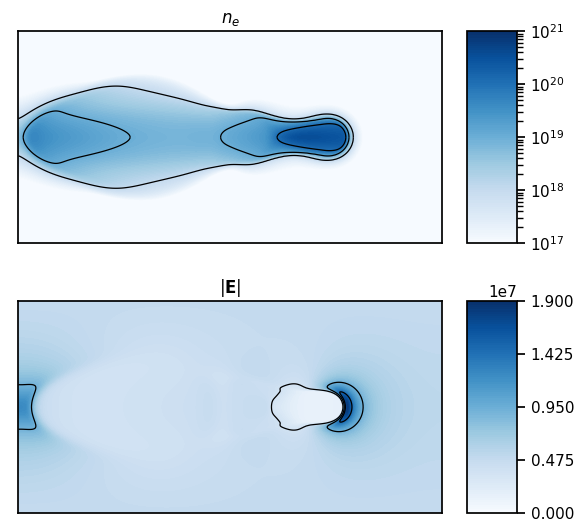}
        \caption{Linear system}
    \end{subfigure}
    \caption{Comparison of electron density and electric field norm at 2.8 ns for neural network and linear system Poisson solver. The computational domain has been mirrored from the central axis.}
    \label{fig:comparison_streamer_nn_vs_ls_28ns}
\end{figure}

The strong background electric field imposed in the whole domain allows ionization of air through collisions and the propagation of two streamers, one going to the left (negative streamer) and the other to the right (positive streamer). The UNet5-$\mrm{RF}_x800$-$\mrm{RF}_y200$ and linear system Poisson solver results are compared in Figs.~\ref{fig:comparison_streamer_nn_vs_ls_16ns} and \ref{fig:comparison_streamer_nn_vs_ls_28ns}. At the beginning of the propagation, the neural network and the linear system yield similar fields (Fig.~\ref{fig:comparison_streamer_nn_vs_ls_16ns}). After a while the absolute values of maximum of electric field and electron density are underestimated by the neural network, where the electric field and electron density profiles are slightly diffused by the network (Fig.~\ref{fig:comparison_streamer_nn_vs_ls_28ns}). Overall, a good agreement is found to be satisfactory as the UNet5 manages to predict correctly the electric field $\vb{E}$, which then drives the propagation of the two streamers.

Two important global properties of these streamers are of interest and need to be well reproduced by the simulations: the speed of the negative and positive streamers as well as the discharge energy. The position of the negative and positive streamers at the axis $r=0$ can be evaluated by the location $x$ of the maximum of the norm of the electric field. The discharge energy $E_d$ is given by \cite{celes2008}

\begin{equation}
    E_d(t) = \int_0^t \int_V \vb{J} \cdot \vb{E} \, \mrm{d} V \, \mrm{d} t
\end{equation}

\noindent where $\vb{J}$ is the total current and the space integration is performed on the entire simulation domain. Since only electrons are moving in this case $\vb{J} = e n_e \mu_e \vb{E}$.

\begin{figure}[htbp]
    \centering
    \includegraphics[width=0.6\textwidth]{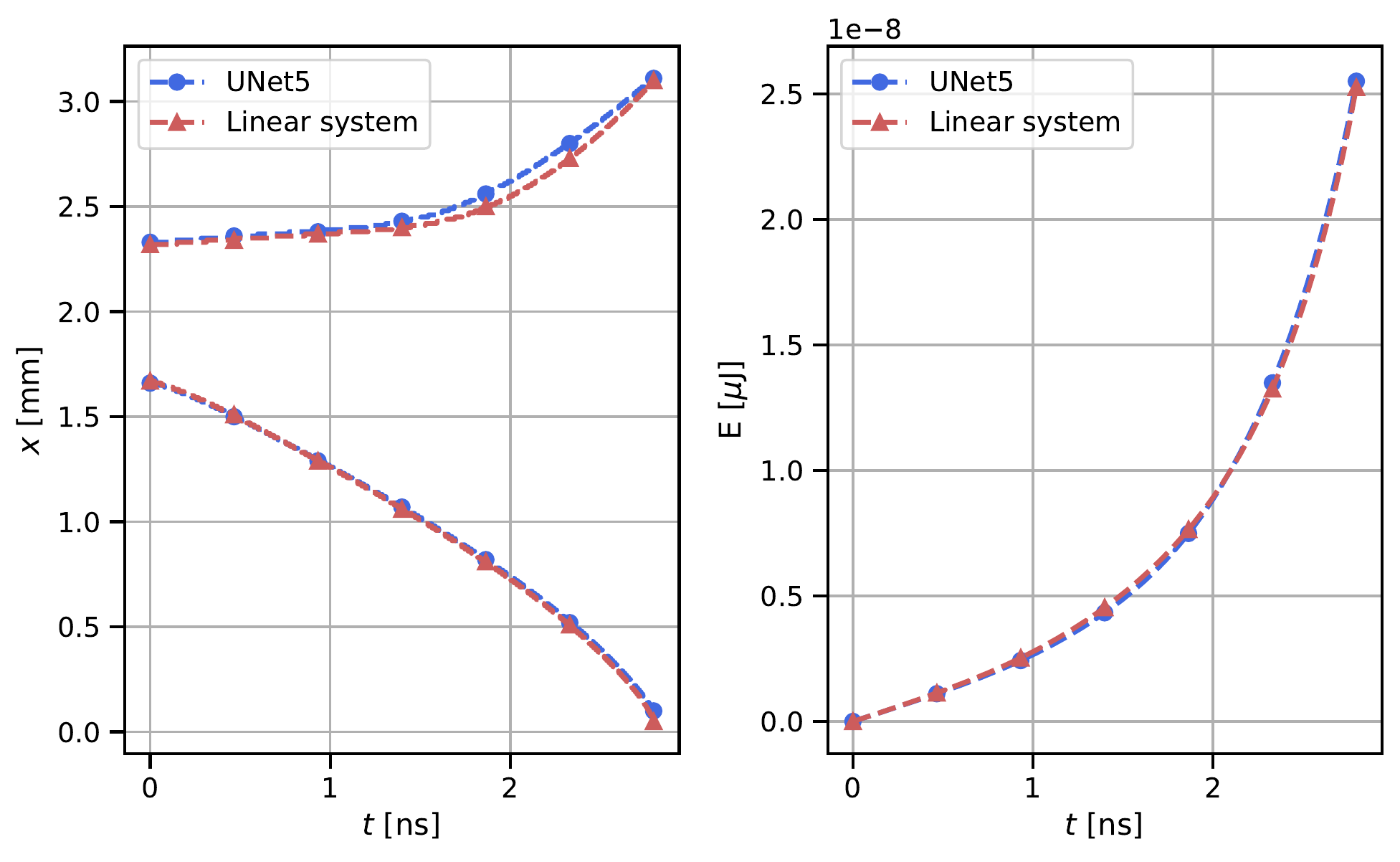}
    \caption{Comparison of positive (above 2 mm) and negative (below 2 mm) streamers propagation and energy for linear system and network runs.}
    \label{fig:streamer_global_comp}
\end{figure}

These quantities are compared in Fig.~\ref{fig:streamer_global_comp}. Although the network does not reproduce exactly the magnitudes of both the electric field and electron density, it can be observed that those global properties are well reproduced by the neural network Poisson solver.

\section{Conclusion}

CNNs have been used to solve the Poisson equation in a 2D-plasma cartesian geometry simulation. Two types of well-known multiple-scale architectures have been used to predict the potential field from the charge density field: UNet and MSNet. These architectures have been chosen in accordance with the analytical solution of the problem at hand which exposes multiple scales.

Across all the hyperparameters varied, the UNet architecture outperforms the MSNet architecture. This has been attributed to the way MSNets encode information from one scale to the other.

The choice of the loss is critical to get a stable simulation. Although the potential is the solved quantity, the real physical value of interest with regard to simulations is the electric field. From the transport equations, $\mathcal{C}^2$ regularity must be ensured and so the \texttt{LaplacianLoss} has been chosen with \texttt{DirichletLoss} to have a reference of potential.

Due to the elliptic nature of the Poisson equation, information propagation across the convolutional layers is critical to get an accurate solution. Increasing the receptive field of the network yields better performance because the fundamental mode is better captured that way. For a fixed receptive field, networks with high number of branches should be preferred as they are faster. However, above a certain number of branches, convolutional layers applied on images with not enough pixels are not understood by the network so that there is an optimum number of branches per resolution.

Generalization of the information from one resolution and one domain length is extended to different ones thanks to scaling laws. Loss of accuracy when applying the network to higher resolutions than the trained one is observed and is amplified for low frequencies.

The performance of neural networks solvers has been assessed and is comparable to that of linear system solvers on the hardware configurations studied. It opens the path to further studies to improve AI-based simulations in an HPC context, for instance to reduce the memory load of deep networks.

The coupling of the neural network in place of a linear solver with plasma transport equations has been tested. The best network found from the inference study is also the most stable one inside the simulation.

Lastly the optimal parameters found in this simple cartesian geometry in terms of receptive field and losses have been adapted to a rectangular domain representing an azimuthal cut of the cylindrical geometry. The methodology adopted for the cartesian, full Dirichlet Poisson problem is shown to be valid also for cylindrical, mixed Dirichlet-Neumann boundary conditions showing a good generalization of the method.

Future works could be dedicated to solve the Poisson equation on unstructured meshes, either for plasma-fluid, or incompressible simulations, for instance to take into account the presence of obstacles (\textit{e.g.} the anode and cathode). Additionnally, this work can be extended to other elliptic equations, such as the screened Poisson equation, which governs photoionisation in plasma-fluid simulations. For all these cases, the insight gained by the present study on the architecture, receptive fields, and performances of neural networks in the context of unsteady simulations could be transferable to these future challenges.

\section*{Acknowledgments}

\noindent This work was supported by the ANR project GECCO (ANR-17-CE06-0019).

\appendix
  \section{Derivation of the analytical solution}
  \label{appendix:analytical_solution}
  The analytical solution of the Poisson equation with boundary conditions depends on the
  Green function $G$ of the chosen configuration \cite[Chap. 1.10]{jackson_electrodynamics}:

  \begin{equation}
      \phi(\vb{x}) = \frac{1}{4\pi\epsilon_0}\int \rho(\vb{x}') \green \dd{V'} + \frac{1}{4\pi} \int \qty(G\pdv{\phi}{n'} - \phi\pdv{G}{n'})\dd{S'}
  \end{equation}

  The Green function of a square domain $L^2$ with zero potential at the four boundaries is \cite[Chap. 3.12]{jackson_electrodynamics}:

  \begin{equation}
      G(x, y, x', y') = \frac{16}{\pi L_x L_y} \sum_{n = 1}^{+\infty}\sum_{m = 1}^{+\infty} \frac{\sin\qty(\frac{n\pi x}{L_x})\sin\qty(\frac{n\pi x'}{L_x})\sin\qty(\frac{m\pi y}{L_y})\sin\qty(\frac{m\pi y'}{L_y})}{n^2/L_x^2 + m^2 / L_y^2}.
  \end{equation}
  In our case $\phi = 0$ on the boundaries and $G = 0$ on the boundaries has to be
  satisfied. Hence:
  \begin{equation}
      \phi(\vb{x}) = \frac{1}{4\pi\epsilon_0}\int \rho(\vb{x}') \green \dd{V'}.
  \end{equation}
  Substituting the Green function, the following solution is obtained:

  \begin{align}
      \phi(x, y) =  \sum_{n = 1}^{+\infty}\sum_{m = 1}^{+\infty} &\qty[\frac{4}{L_x L_y} \int_{x', y'} \sin\qty(\frac{n\pi x'}{L_x})\sin\qty(\frac{m\pi y'}{L_y})R(x', y') \dd{x'} \dd{y'}] \nonumber \\
      &\times \frac{\sin\qty(\frac{n\pi x}{L_x})\sin\qty(\frac{m\pi y}{L_y})}{\pi^2(n^2/L_x^2 + m^2 / L_y^2)}. \label{eq:appendix_analytical_solution}
  \end{align}

  \section{Normalization of inputs}
  \label{appendix:normalization}

  A reasonable value for the value of the ratio of the potential over the charge density needs to be found. From
  the solution of the potential in terms of Fourier series Eq.~\eqref{eq:appendix_analytical_solution}, assuming a constant potential and taking only the term in $n = 1, m = 1$ in the summation reduces to:

  \begin{align}
      \phi(x, y) = \frac{4}{L_x L_y} \int_{x', y'} \sin(\frac{n\pi x'}{L_x})\sin(\frac{m\pi y'}{L_y})R(x', y') \dd{x'} \dd{y'} \times \frac{\sin(\frac{\pi x}{L_x})\sin(\frac{\pi y}{L_y})}{\pi^2 /L_x^2 + \pi^2 / L_y^2}.
  \end{align}
  Taking the absolute value, the following inequality holds:
  \begin{equation}
      \qty|\frac{\phi}{R}|_\text{max} \leq \frac{1}{\qty(\frac{\pi^2}{4})^2\qty(\frac{1}{L_x^2} + \frac{1}{L_y^2})}.
  \end{equation}
  Therefore:
  \begin{equation}
      \qty|\frac{\phi}{R}|_\text{max} = \frac{\alpha}{\qty(\frac{\pi^2}{4})^2\qty(\frac{1}{L_x^2} + \frac{1}{L_y^2})}
  \end{equation}
  with $\alpha \leq 1$.

  \section{Linear system solvers benchmark using PETSc}
  \label{appendix:petsc_benchmark}
\begin{figure}[htbp]
      \centering
      \begin{subfigure}[b]{0.45\textwidth}
          \centering
          \includegraphics[width=\textwidth]{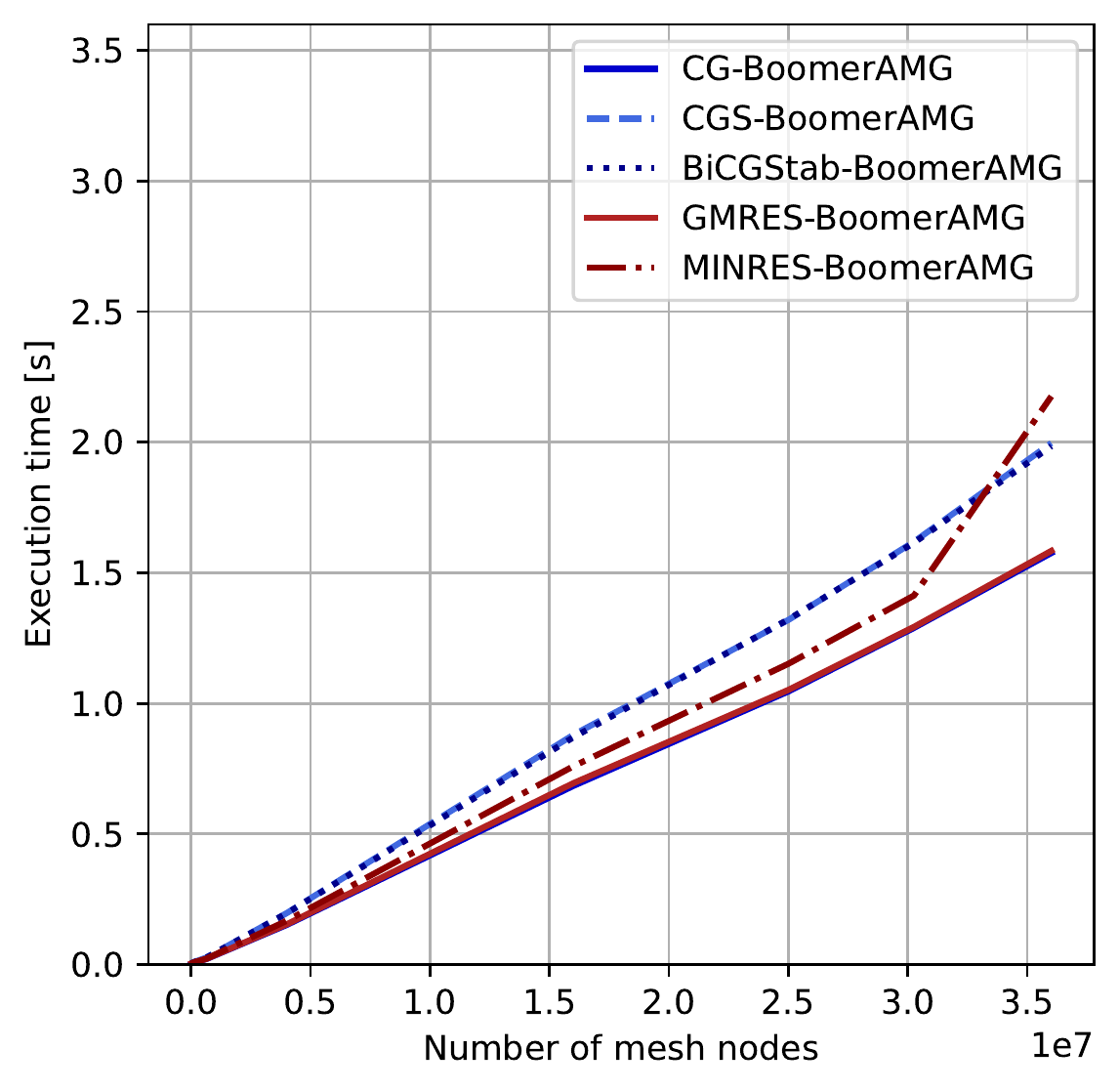}
          \caption{BoomerAMG}
          \label{fig:config_1_boomeramg}
      \end{subfigure}
      \begin{subfigure}[b]{0.45\textwidth}
          \centering
          \includegraphics[width=\textwidth]{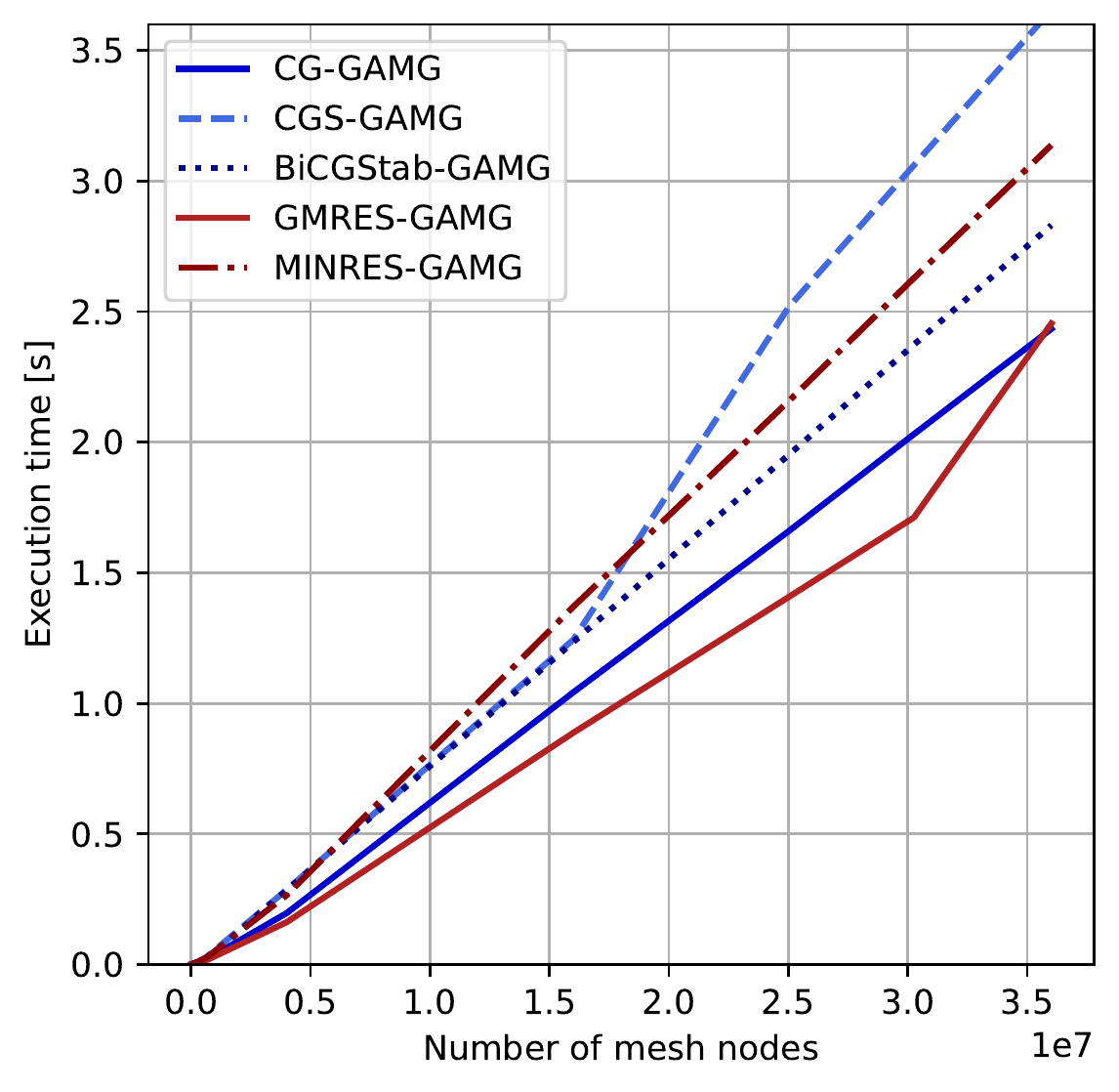}
          \caption{PETSc GAMG}
          \label{fig:config_1_gamg}
      \end{subfigure}
      \caption{Performance of different solvers using AMG preconditioning on \texttt{config\_1}.}
      \label{fig:solvers_config_1}
  \end{figure}

  A varierty of solvers have been benchmarked in PETSc and are shown in Fig.~\ref{fig:solvers_config_1}. The most popular and used iterative solvers have been tested: Conjugate Gradient (CG), Conjugate Gradient Squared (CGS), Stabilized Biconjugate Gradient (BiCGStab), Minimal Residual (MINRES) and Generalized Minimal Residual (GMRES). The preconditioner is critical to get good performance and since the problem is elliptic, multigrid preconditioners are best suited for them \cite{amg_review}. The native PETSc GAMG \cite{petsc-web-page} and Hypre BoomerAMG \cite{hypre_ref} resolution times in combination with the aforementioned iterative solvers are shown in Figs.~\ref{fig:config_1_boomeramg} and \ref{fig:config_1_gamg}, respectively. Other preconditioners have been tested but due to huge performance gap compared to the multi-grid preconditioners they are not shown. BoomerAMG is shown to outperform the native PETSc GAMG for every iterative solver used. When using BoomerAMG, GMRES and CG yield very similar results with a small edge for CG which has been retained in Fig.~\ref{fig:ls_vs_nn_perfs}.

\bibliographystyle{unsrt}
\bibliography{refs}
\clearpage

\end{document}